 \documentclass[acmsmall, screen]{acmart}

\newtheorem{theorem}{Theorem}[section]

\newtheorem{prop}[theorem]{Proposition}
\newtheorem{assumption}[theorem]{Assumption}

\newtheorem{rem}[theorem]{Remark}
\newtheorem{ex}[theorem]{Example}

\usepackage{mathtools}

\usepackage{algorithm}
\usepackage[noend]{algpseudocode}
\usepackage{multirow}
\usepackage{mysymbol}
\usepackage[title,titletoc]{appendix}
\usepackage{wrapfig}
\usepackage{booktabs}

\AtBeginDocument{%
  }

\setcopyright{acmlicensed}
\acmJournal{TCPS}

\begin{document}

\title{Conformal Temporal Logic Planning using Large Language Models}

\author{Jun Wang}
\orcid{0000-0003-0481-8697}
\authornote{Corresponding author - ioannisk@wustl.edu. This work was done when the third author was at Washington University in St. Louis. This work was supported by the NSF award CNS \#2231257. \textcolor{black}{Additional materials, including videos, are available on the project webpage: \href{https://ltl-llm.github.io/}{\textbf{https://ltl-llm.github.io}}}}
\affiliation{%
  \institution{Washington University in St Louis}
  \city{St Louis}
  \state{MO}
  \country{USA}}

\author{Jiaming Tong}
\orcid{0009-0009-7423-7311}
\affiliation{%
  \institution{University of Zurich}
  \city{Zurich}
  \country{Switzerland}}

\author{Kaiyuan Tan}
\orcid{0009-0003-6069-813X}
\affiliation{%
  \institution{Vanderbilt University}
  \city{Nashville}
  \state{TN}
  \country{USA}
}

\author{Yevgeniy Vorobeychik}
\orcid{0000-0003-2471-5345}
\affiliation{%
 \institution{Washington University in St Louis}
 \city{St Louis}
  \state{MO}
  \country{USA}
  }

\author{Yiannis Kantaros}
\orcid{0009-0003-6081-5092}
\affiliation{%
  \institution{Washington University in St Louis}
  \city{St Louis}
  \state{MO}
  \country{USA}}

\begin{abstract}
This paper addresses temporal logic task planning problems for mobile robots. We consider missions that require accomplishing multiple high-level sub-tasks, expressed in natural language (NL), in a temporal and logical order. To formally define the mission, we treat these sub-tasks as atomic predicates in a Linear Temporal Logic (LTL) formula. We refer to this task specification framework as LTL-NL. Our goal is to design plans, defined as sequences of robot actions, accomplishing LTL-NL tasks. This action planning problem cannot be solved directly by existing LTL planners due to the NL nature of atomic predicates. Therefore, we propose HERACLEs, a hierarchical neuro-symbolic planner that relies on a novel integration of (i) existing symbolic planners generating high-level task plans determining the order at which the NL sub-tasks should be accomplished; (ii) pre-trained Large Language Models (LLMs) to design sequences of robot actions for each sub-task in these task plans; and (iii) conformal prediction acting as a formal interface between (i) and (ii) and managing uncertainties due to LLM imperfections. We show, both theoretically and empirically, that HERACLEs can achieve user-defined mission success rates.  We demonstrate the efficiency of HERACLEs through comparative numerical experiments against recent LLM-based planners as well as hardware experiments on mobile manipulation tasks. Finally, we present examples demonstrating that our approach enhances user-friendliness compared to conventional symbolic approaches.
\end{abstract}



\maketitle

\vspace{-0.3cm}
\section{Introduction}\label{sec:intro}
\vspace{-0.1cm}


Designing autonomous agents with task planning capabilities is a long-standing goal in robotics and automation \cite{garrett2021integrated,antonyshyn2023multiple}. Achieving this goal requires the development of mission specification frameworks and planning algorithms capable of generating plans—sequences of robot actions—to accomplish missions. These frameworks should allow users to define missions unambiguously and in a user-friendly manner, while planners need to exhibit computational efficiency and be supported by correctness guarantees.

Recently, several task planners have been proposed that can design correct-by-construction plans for complex missions with specified temporal and logical requirements using Linear Temporal Logic (LTL) \cite{sahin2019multirobot,kantaros2022perception,luo2021abstraction,ulusoy2014optimal,sun2022neurosymbolic,kantaros2020stylus,vasile2013sampling,ho2023planning,zhou2023vision,kantaros2020reactive,vasilopoulos2021reactive,chen2024distributed,leahy2022fast,guo2015bottom,shoukry2017linear,hashimoto2022collaborative, kantaros2017sampling, chen2018verifiable}.
However, this framework demands a significant amount of expertise and manual effort to rigorously define complex tasks. This is because defining LTL tasks requires specifying multiple atomic predicates (i.e., Boolean variables) to model desired low-level robot configurations and coupling them using temporal/Boolean operators. Additionally, the more complex the task requirements, the larger the number of predicates and temporal/logical operators needed to define the corresponding LTL formula. This not only compromises the user-friendliness but also increases the computational cost of designing robot plans \cite{kantaros2020stylus}.
%
%
%
%
%
On the other hand, Natural Language (NL) has emerged as a more user-friendly method for specifying robot missions. Early research in NL-based planning mainly focused on mapping NL to planning primitives \cite{tellex2011understanding,matuszek2010following,chen2011learning,howard2014natural} using statistical machine translation \cite{koehn2009statistical} to identify data-driven patterns for translating free-form commands into a formal language defined by a grammar.  However, these approaches were limited to structured state spaces and simple NL commands. Motivated by the remarkable generalization abilities of pre-trained Large Language Models (LLMs) across diverse task domains there has been increasing attention on utilizing LLMs for NL-based planning 
\cite{singh2023progprompt,ren2023robots, liang2023code,shah2023lm,ding2023task,liu2023llm+,stepputtis2020language, li2022pre,huang2022inner,ahn2022can,ruan2023tptu,luo2023obtaining,joublin2023copal,yang2024text2reaction,rana2023sayplan,wang2024llm}. While LLM-based planning methods are more user-friendly than LTL-based planners, they have two main limitations: (i) they lack mission performance guarantees; and (ii) their ability to design correct plans deteriorates with increasing task complexity \cite{chen2023autotamp}. \textcolor{black}{In this paper, we propose a new approach to \textit{mission specification} and \textit{robot planning} that combines the formal rigor of LTL-based planners with the user-friendliness of NL-based approaches.} 

\textcolor{black}{First, we discuss our proposed \textit{mission specification framework}.
Our specification framework, called LTL-NL, models complex tasks as collections of NL-described sub-tasks with logical and temporal dependencies expressed using standard Boolean and temporal operators. Each NL sub-task is treated as an atomic predicate—evaluated as true when successfully completed, and false otherwise. The key departure from conventional LTL-based methods lies in specifying predicates/sub-tasks using NL, at the level of intent, rather than low-level system states (see, e.g., \cite{schuppe2020multi,chen2012formal,kantaros2020stylus,vasile2013sampling,he2015towards}). For instance, consider a task where a robot must first deliver a bottle of water to the kitchen table, and then deliver a Coke to the office desk. We can define two NL sub-tasks as predicates, `deliver a bottle of water to the kitchen table' and `deliver a Coke to the office desk', and use Boolean/temporal operators to express the logical and temporal relationship between them. In contrast, conventional LTL formulas require low-level predicates that are true if e.g., the robot is near the bottle or if it is holding the object. Our LTL-NL framework results in a more compact and user-friendly mission specification compared to conventional LTL formulas.
In comparison to NL instructions, 
our framework enables the automatic decomposition of the overall mission (i.e., an LTL-NL specification) into multiple NL-based sub-tasks using existing symbolic planners \cite{kantaros2020reactive,vasilopoulos2021reactive,leahy2022fast}.  It is important to note that such a decomposition is highly challenging to perform in a correct-by-construction fashion for missions expressed exclusively using NL. This decomposition is critical for ensuring logical consistency and robust performance -particularly as mission complexity increases- enabling our planning framework to consistently outperform existing LLM-based planners.}

\begin{figure*}[t] 
\centering
\includegraphics[width=\linewidth]{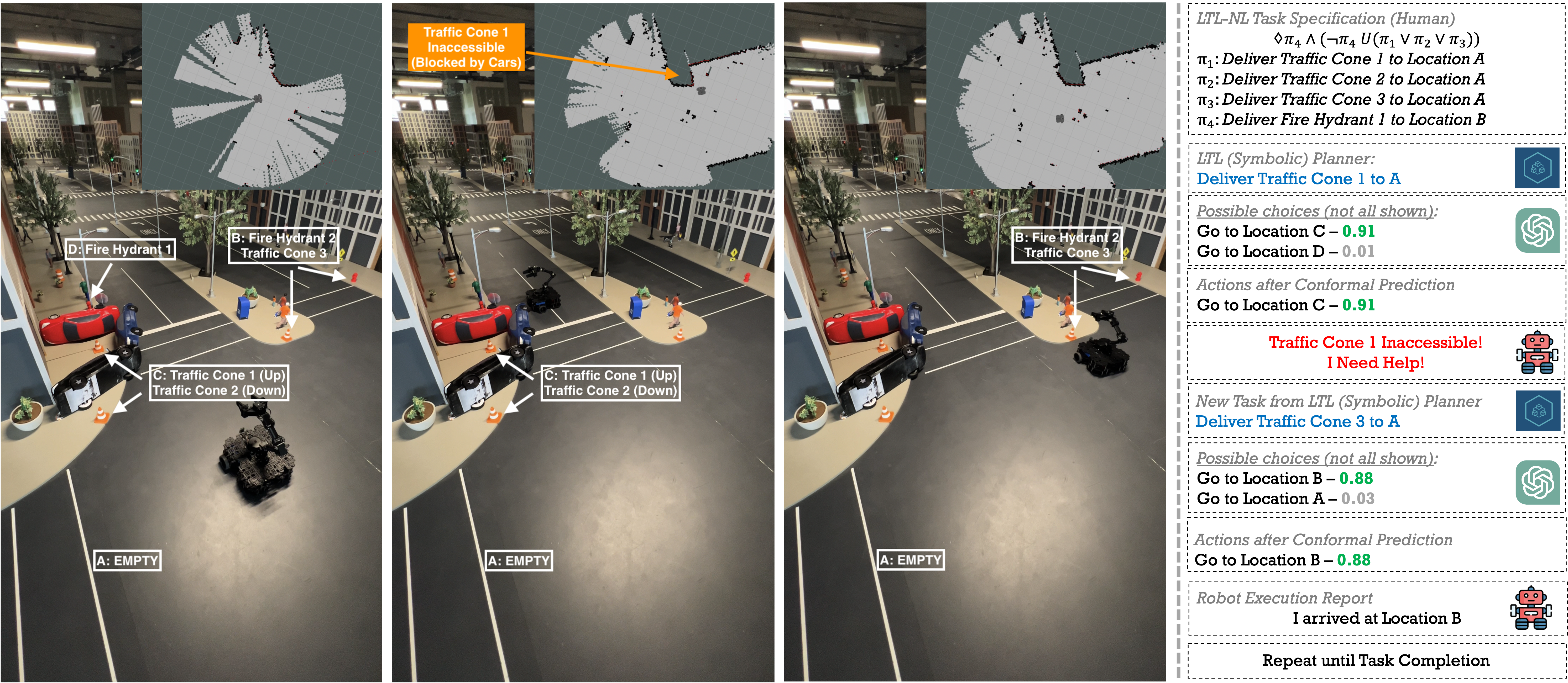} \vspace{-0.7cm}
\caption{
\textcolor{black}{This figure illustrates a robot operating in a post-disaster city. The high-level mission-"deliver either traffic cone 1, 2, or 3 to location A, and only then deliver fire hydrant 1 to location B"-is formulated with an LTL-NL formula $\phi=\diamond \pi_4 \wedge (\neg \pi_4 \ccalU (\pi_1 \vee \pi_2 \vee \pi_3))$ with 4 NL subtasks ($\pi_1\sim \pi_4$). The symbolic planner initially generates the NL-subtask $\pi_1$ (`deliver traffic cone 1 to location A') \textcolor{black}{along with the safety constraint $\neg \pi_4$ (`do not move the fire hydrant 1 to location B')}. As the robot executes the LLM-generated plan to accomplish this sub-task, it builds a grid map of the environment revealing that traffic cone 1 is blocked by surrounding cars, making it inaccessible. Consequently, the symbolic planner assigns a new sub-task, \textcolor{black}{modeled by $\pi_3$ and $\neg \pi_4$}, requiring the robot to deliver traffic cone 3 to A; see also Example \ref{ex:me} and the demonstrations in \cite{ltl_llm_video_no_name}.}
%
%
} 
\label{fig:env} \vspace{-0.5cm}
\end{figure*}

\textcolor{black}{Second, we address the challenge of \textit{designing robot plans} that complete LTL-NL tasks.} Specifically, we consider a mobile robot equipped with various skills (e.g., mobility, manipulation, and sensing) tasked with missions expressed as LTL-NL specifications. Each predicate in the LTL-NL formula is satisfied if an NL-based sub-task is accomplished, requiring the robot to apply its skills to various semantic objects and regions of interest in the environment. For instance, the sub-task `deliver a bottle of water to the kitchen table' requires the robot to move to the location where the bottle is, grasp it, move to the kitchen table, and release the object. Observe that a predicate in an LTL-NL formula is satisfied by a sequence of robot configurations/actions.
While we assume that the locations and semantic labels of these objects are known, the geometric environmental structure is initially unknown. The latter may hinder the completion of certain NL-based sub-tasks (e.g., when an object is inaccessible); see Fig. \ref{fig:env}. Our goal is to design robot plans, defined as a sequence of robot actions, satisfying the assigned LTL-NL missions. 

\textcolor{black}{
%
%
To address this planning problem, we propose a hierarchical neuro-symbolic planner consisting of (1) a \textit{symbolic planner}, determining the next NL-encoded sub-task the robot should execute to make mission progress, and (2) a \textit{neural planner}, which translates these sub-tasks into executable robot actions.
In our implementation, we adopt the symbolic planner from \cite{kantaros2020reactive,vasilopoulos2021reactive} due to its computational efficiency, though our framework remains compatible with other symbolic planners.
Given the NL sub-task selected by the symbolic planner, the neural planner uses a pre-trained LLM to generate a robot plan that satisfies it \cite{touvron2023llama,brown2020language}.}
\textcolor{black}{It is possible that the initially unknown geometric structure of the environment may prevent the robot from fully executing a generated plan (e.g., due to obstacles blocking access to regions of interest). In such cases, a `help' message is sent to the symbolic planner, which responds by issuing alternative sub-tasks to the neural planner, if available, \cite{kantaros2020reactive,vasilopoulos2021reactive}; see Fig.~\ref{fig:env}. Once a sub-task is completed, the symbolic planner selects the next one and forwards it to the neural planner for plan generation.
This interaction between the symbolic and neural planners continues until the overall LTL-NL mission is completed.}
A key challenge in implementing the neural planner is that LLMs tend to hallucinate, i.e., to confidently generate incorrect outputs \cite{lecun2023grounding}. 
To reason about the correctness of the LLM-generated plans, inspired by \cite{ren2023robots,kumar2023conformal}, we leverage conformal prediction (CP) \cite{angelopoulos2021gentle}, a statistical tool for uncertainty quantification in black-box models that has been recently applied to various safe autonomy tasks \cite{angelopoulos2023conformal,kumar2023conformal,ren2023robots,mao2023safe,muthali2023multi,tonkens2023scalable,lindemann2023conformal,yang2024memory,qin2024statistical}.
CP constructs on-the-fly prediction sets that contain the correct robot action with user-specified confidence. This allows the LLM to determine when it is uncertain about its predictions. In cases of high uncertainty, indicated by non-singleton prediction sets, \textcolor{black}{the LLM seeks assistance from a user or the symbolic planner.} 
%
%
\textcolor{black}{This formal interface between the symbolic planner and the LLM, enabled by CP,} gives rise to our planner, called HERACLEs, for HiERArchical ConformaL natural languagE planner. The generated plans are executed using existing low-level controllers. \textcolor{black}{We show both theoretically and empirically that HERACLEs can achieve user-specified mission completion rates, thanks to CP, while requiring low help rates—significantly lower than related works \cite{ren2023robots}, due to its neuro-symbolic architecture.}
%
%
%
%

\textbf{Related works:} \textbf{(i)} As discussed earlier, a substantial body of \textit{symbolic planners} exists that can design plans satisfying LTL tasks \cite{sahin2019multirobot,kantaros2022perception,luo2021abstraction,ulusoy2014optimal,sun2022neurosymbolic,kantaros2020stylus,vasile2013sampling,ho2023planning,zhou2023vision,kantaros2020reactive,vasilopoulos2021reactive,chen2024distributed,leahy2022fast,guo2015bottom,shoukry2017linear, hashimoto2022collaborative, kantaros2017sampling, chen2018verifiable}.
%
We emphasize that these planners cannot solve the considered LTL-NL planning problem due to the NL nature of the predicates. Specifically, these works can design a sequence of NL predicates that need to be satisfied to accomplish the LTL-NL mission; however, they cannot design a sequence of robot actions (e.g.,  `grab the bottle of water at location A', `go to the kitchen table', `drop off the bottle') that satisfies a NL-based predicate (e.g., `deliver a bottle of water to the kitchen table'), let alone an LTL-NL formula, as they cannot process NL. We note again that our approach is tightly integrated with these symbolic planners as they are used to generate a sequence of NL sub-tasks that should be accomplished to satisfy an LTL-NL formula.
\textbf{(ii)} Related are also recent \textit{neural planners} that can design plans for tasks, described solely in NL, using pre-trained LLMs; see e.g., \cite{singh2023progprompt,ren2023robots, liang2023code,shah2023lm,ding2023task,liu2023llm+,stepputtis2020language, li2022pre,huang2022inner,ahn2022can,ruan2023tptu,luo2023obtaining,joublin2023copal,yang2024text2reaction,rana2023sayplan,wang2024llm}; a recent survey can be found in \cite{pallagani2024prospects,hunt2024survey}. However, the ability of LLMs to design correct plans drops significantly as the mission complexity increases as shown in \cite{chen2023autotamp,valmeekam2024planning} as well as in our experiments.  
%
Additionally, as discussed earlier, the majority of these planners lack correctness and mission completion guarantees. \textcolor{black}{Among related works, the most closely aligned is KnowNo \cite{ren2023robots}, which also employs CP to quantify uncertainty in LLM outputs and enables robots to seek help from users to achieve desired mission completion rates. However, as demonstrated in our comparative experiments, HERACLEs achieves significantly lower help rates, primarily due to its task decomposition enabled by the symbolic planner. Moreover, HERACLEs can identify specific sub-tasks that cannot be completed due to the geometric constraints of the environment and adapt by requesting alternative sub-tasks from the symbolic planner. This reactive capability is not supported by \cite{ren2023robots}, and as a result, it cannot readily adapt to unknown environments.}
%
\textbf{(iii)} \textit{Neuro-symbolic planners} that integrate LLMs with symbolic planners have also been proposed recently in \cite{chen2023autotamp,quartey2024verifiably,dagan2023dynamic,fuggitti2023nl2ltl,chen2023nl2tl,cosler2023nl2spec,dai2023optimal}. Their key idea is to employ LLMs to translate NL commands into temporal logic specifications. Once this translation is completed, existing symbolic planners can be employed to design correct plans. A key limitation is that the translation process is not supported by correctness guarantees. As a result, the resulting plan may not satisfy the original NL instructions. In contrast, HERACLEs employs LLMs and symbolic planners in a fundamentally different way while CP allows the design of plans that are probabilistically correct and, therefore, they can achieve desired mission completion rates. 
%


%
\textbf{Comparative Evaluations:} 
%
We provide extensive comparative experiments to demonstrate the benefits of the proposed method. First, we compare  HERACLEs against KnowNo \cite{ren2023robots} as it is the closest work to ours. As discussed earlier, KnowNo can also achieve desired mission success rates by incorporating CP in the plan synthesis phase and allowing robots to ask for help in cases of high uncertainty. Since KnowNo requires the mission to be expressed in NL, as opposed to LTL-NL, in our comparisons, we manually translate LTL-NL missions into NL instructions serving as inputs to \cite{ren2023robots}. 
\textcolor{black}{Our empirical analysis demonstrates that HERACLEs achieves significantly lower help rates than \cite{ren2023robots} to attain the same mission success rate. We attribute the performance gap between KnowNo and HERACLEs primarily to the LTL-NL framework and the symbolic planner decomposing missions into multiple smaller/simpler NL-based sub-tasks that LLMs can handle more efficiently. }
Second, we provide examples demonstrating the user-friendliness of LTL-NL formulas compared to corresponding LTL specifications. \textcolor{black}{Specifically, we show that LTL-NL formulas require significantly fewer predicates than LTL formulas defined over low-level configuration-based predicates modeling the same task.} This difference in the `length' of LTL and LTL-NL formulas becomes more pronounced as the complexity of the mission
requirements increases. Here we would like to note that direct comparisons with other temporal logic planners are not possible, as they cannot handle LTL-NL formulas, as discussed earlier. Similarly, comparisons with the neuro-symbolic approaches mentioned above are not straightforward, since they address NL-to-LTL translation problems, while our focus is on solving planning problems once an LTL-NL formula is specified. In Appendix \textcolor{magenta}{\ref{sec:CompI}}, we also provide additional comparisons against the LLM-based planners proposed in \cite{singh2023progprompt,ahn2022can,chen2023scalable}, which, however, do not allow robots to ask for help and, therefore, they cannot achieve desired mission completion rates. These comparative experiments show that the ability of these approaches to design correct plans tends to decrease as the number of temporal and logical requirements in the mission increases. To the contrary, the planning performance of HERACLEs consistently outperforms these baselines (even when the help mode is deactivated).

\textbf{Contribution:} The contribution of the paper can be summarized as follows. \textit{First}, we propose a new \textcolor{black}{mission specification framework},  called LTL-NL, to define complex high-level robot tasks. The advantage of LTL-NL over conventional LTL is its user-friendliness, as it requires fewer predicates to define the same task while the predicates can be defined using NL rather than low-level system configurations.
\textit{Second}, we introduce HERACLEs, a new neuro-symbolic planner that integrates symbolic planners with LLMs to design plans satisfying missions encoded as LTL-NL formulas \textcolor{black}{in partially known environments}. \textit{Third}, we show, both theoretically and empirically, that HERACLEs can achieve user-specified mission success rates through a formal interface, employing CP, between the LLM and the symbolic planner. \textit{Fourth}, \textcolor{black}{we conduct extensive comparative experiments showing that HERACLEs outperforms state-of-the-art LLM-based planners both in generating correct plans and in minimizing human intervention (help rates).}
\textcolor{black}{\textit{Finally}, we provide hardware experiments on mobile manipulation tasks to demonstrate the effectiveness of our method.} 

\vspace{-0.2cm}
\section{Problem Formulation}\label{sec:problem}
\textbf{Robot System and Skills:} Consider a robot governed by the following deterministic and known dynamics: $\bbp(t+1)=\bbf(\bbp(t),\bbu(t))$, 
where $\bbp(t)$ stands for the robot state (e.g., position and orientation), and $\bbu(t)$ stands for control input at discrete time $t$. 
%
The robot has $A>0$ number of abilities/skills collected in a set $\ccalA\in\{1,\dots,A\}$. Each skill $a\in\ccalA$ is represented as text such as `take a picture', `grab', or `move to'. Application of a skill $a$ at an object/region with location $\bbx$ at time $t\geq0$ is denoted by $s(a,\bbx,t)$ or, for brevity, when it is clear from the context, by $s(t)$. 
The time step $t$ is increased by one, once an action is completed. We assume that the robot has access to low-level controllers to apply the skills in $\ccalA$.  

\textbf{Partially Known Semantic Environment:} The robot operates within a semantic environment $\Omega\subseteq\mathbb{R}^d$, $d\in\{2,3\}$ with fixed, static, and potentially unknown obstacle-free space denoted by $\Omega_{\text{free}}\subseteq\Omega$. The space $\Omega_{\text{free}}$ is populated with $M>0$ static semantic objects. Each object $e$ is characterized by its location $\bbx_e$ and semantic label $o_e$ (e.g., `bottle' or `chair').
The robot is assumed to have knowledge of both the location and label of each object. Objects may also be located inside containers (e.g., drawer or fridge), with their status (open/closed) initially unknown. Also, the occupied space $\Omega\setminus\Omega_{\text{free}}$ may prevent access to certain semantic objects rendering execution of certain decisions $s(t)$ infeasible. We assume that the robot is equipped with sensors allowing it to detect obstacles and reason about the status of containers containing objects of interest. Hereafter, we denote by $\ccalS$ a finite set collecting all decisions $s(t)$ the robot can take. 
\textcolor{black}{This set is constructed offline using the action space $\ccalA$ and the corresponding available objects.}


\textbf{Mission Specification:} The robot is tasked with a high-level mission with temporal and logical requirements. To formally define such tasks, we employ Linear Temporal Logic (LTL). LTL is a formal language that comprises a set of atomic propositions (AP) (i.e., Boolean variables), denoted by $\mathcal{AP}$, Boolean operators, (i.e., conjunction $\wedge$, and negation $\neg$), and temporal operators, such as \textit{always} $\square$, \textit{eventually} $\lozenge$, and until $\mathcal{U}$; see also Example \ref{ex:me}. 
A formal presentation of the syntax and semantics of LTL can be found in \cite{baier2008principles}. For simplicity, hereafter, we restrict our attention to co-safe LTL formulas that are a fragment of LTL and that exclude the `always' operator. \textcolor{black}{Such} 
formulas can be satisfied within a finite horizon $H$. 
\textcolor{black}{Throughout the paper, for brevity,  we will use the term LTL instead of co-safe LTL. }
We define APs so that they are true when a sub-task expressed in natural language (NL) is satisfied, and false otherwise. Valid definitions of APs include sub-tasks that can be accomplished by a finite robot trajectory $\tau$, defined as a finite sequence of $T$ decisions selected from $\ccalS$, i.e., 
\begin{equation}\label{eq:tau}
    \tau=s(t),s(t+1),\dots,s(t+k),\dots,s(t+T-1),
\end{equation}
where $k\in\{0,\dots,T-1\}$, for some $T\geq 1$; see also Example \ref{ex:me} and Section \ref{sec:taskspec}. 
We call formulas constructed in this way as co-safe LTL-NL formulas. Co-safe LTL-NL formulas are satisfied by finite-horizon robot trajectories $\tau_{\phi}$ defined as 
\begin{equation}\label{eq:tauphi}
\tau_{\phi}=\tau_1,\dots,\tau_n,\dots,\tau_N,
\end{equation}
where $\tau_n$ is a finite robot trajectory of horizon $T_n$, as defined in \eqref{eq:tau}. Thus, the total horizon $H$ of the plan $\tau_{\phi}$ is $H=\sum_{n=1}^NT_n$.
We highlight that in $\tau_{\phi}$, the index $n$ is different from the time instants $t\in\{1,\dots,H\}$. 
In fact, $n\in\{1,\dots,N\}$ is an index, initialized as $n=1$ and increased by $1$ every $T_n$ time instants, pointing to the next finite trajectory in $\tau_{\phi}$. 

\textbf{Distribution of Mission Scenarios:} \textcolor{black}{We formalize the above by defining a mission scenario as $\xi_i=\{\ccalA_i, \phi_i, H_i, \Omega_i\}$. Recall that $\ccalA_i, \phi_i, H_i, \text{~and~} \Omega_i$ refer to the robot skills, the LTL-NL mission, the mission horizon, and the semantic environment, respectively. The subscript $i$ is used to emphasize that these parameters can vary across scenarios. When it is clear from context, we drop the dependence on $i$. 
We assume that all scenarios are sampled from a distribution $\ccalD$ defined over feasible scenarios $\xi$. 
Note that $\ccalD$ is unknown, but we assume that we can sample i.i.d.~scenarios from it as e.g., in \cite{ren2023robots}.
}

\textbf{Problem Statement:} 
\textcolor{black}{Our goal is to design
safe LTL-NL planners, i.e., planners that can generate correct plans $\tau_{\phi_i}$ satisfying mission scenarios $\xi_i$ in at least $(1-\alpha)\%$ of the scenarios drawn from $\ccalD$ for a user-specified $\alpha\in(0,1)$} under the following assumption \textcolor{black}{(see Rem. \ref{rem:as})}:
\begin{assumption}[Error-free Skills]\label{as:A1}
   The robot has access to low-level controllers, allowing it to apply its skills error-free; \textcolor{black}{see Appendix \ref{sec:relaxPerfectSkills} for a discussion on how this assumption can be relaxed.}
\end{assumption}


\begin{rem}[\textcolor{black}{Error-Free Skills} \& Probabilistic Task Satisfaction]\label{rem:as}
  \textcolor{black}{To address this problem, we will employ LLMs to process the NL-based predicates. The probabilistic satisfaction of a mission arises exclusively due to imperfections of the employed LLM and not due to low-level uncertainties in the execution of robot skills. The latter is precluded by Assumption \ref{as:A1}. 
  We note that Assumption \ref{as:A1} is common in related temporal logic (e.g., \cite{kantaros2020stylus,kamale2023cautious}) and language-based (e.g., \cite{chen2023scalable,ren2023robots}) planners that also aim to design semantically correct plans that, when executed properly, fulfill the task requirements.
  }
%
%
  Formally relaxing this assumption and extending the proposed planner to account for uncertainty in the system dynamics and the performance of the robot skills is part of our future work; \textcolor{black}{see Appendix  \ref{sec:relaxPerfectSkills}.} 
\end{rem}

\begin{ex}[LTL-NL Mission]
Consider a robot with skills $\ccalA=\{\text{go to}, \text{pick up}, \text{put down}\}$. The robot is situated in a post-disaster city where there are $M=6$ objects of interest with $\ccalO=\{\text{Fire Hydrant}, \\  \text{Traffic Cone}\}$.  The environment along with the locations of all semantic objects is shown in Fig. \ref{fig:env}. 
The task of the robot is modeled as an LTL-NL formula $\phi = \Diamond \pi_4 \wedge (\neg  \pi_4 \ccalU (\pi_1 \vee \pi_2 \vee \pi_3))$,
where $\pi_1$,  $\pi_2$, $\pi_3$, $\pi_4$ model the sub-tasks `Deliver Cone 1 to $\bbx_A$', `Deliver Cone 2 to $\bbx_A$', `Deliver Cone 3 to $\bbx_A$', and `Deliver Fire Hydrant 1 to $\bbx_B$', respectively. This formula requires eventually satisfying $\pi_4$ but only after one of the sub-tasks $\pi_1$, $\pi_2$, and $\pi_3$ is satisfied.
Thus, this task can be satisfied by first delivering one of the traffic cones to $\bbx_A$ and then delivering the fire hydrant to $\bbx_B$. 
A plan $\tau_{\phi}$ to satisfy $\phi$ is defined as $\tau_{\phi}=s(\text{go to}, \bbx_B,1),s(\text{pick up}, \text{Cone 3},2), s(\text{go to}, \bbx_A,3), s(\text{put down}, 4), s(\text{go to}, \bbx_D,5), \\s(\text{pick up}, \text{Fire Hydrant 1},6), s(\text{go to}, \bbx_B,7), s(\text{put down}, 8)$; see the demonstration in \cite{ltl_llm_video_no_name}. 
\label{ex:me}
\end{ex}

\vspace{-0.1cm}
\section{Proposed Solution: HiERArchical ConformaL natural languagE planner (HERACLEs)
}\label{sec:method}
\vspace{-0.1cm}

\begin{algorithm}[t]
\footnotesize
\caption{HERACLEs: A Hierarchical Neuro-Symbolic Planner 
}\label{alg1:cp_ltl_planner}
\begin{algorithmic}[1]
\State \textbf{Input}: LTL-NL Task $\phi$; Coverage level $\alpha$; 
\State \textcolor{black}{Initialize $\Sigma_{\text{unc}}(0)=\emptyset$; }\textcolor{black}{LTL planner generates NL-based sub-task ($\pi_{\text{next}}$, $\Sigma_{\text{unsafe}}$)}\label{algo1:dfa}
\State Convert sub-task into an initial prompt $\ell(1)$ with empty history of actions\;\label{algo1:prompt_init}  
\While{($\phi$ not accomplished) $\wedge ~(\pi_{\text{next}}\neq\varnothing)$}
\State $\tau=\varnothing$\label{algo1:nullAction}\;
\For{$k=0~\text{to}~T-1$} \label{algo1:label_k}
\State Compute prediction set $\mathcal{C}(\ell(t+k))$ (Alg. \ref{alg2:alg2})\label{algo1:pred_set}
\If{ $|\mathcal{C}(\ell(t+k))|>1$ } \Comment{\textcolor{black}{Ask-for-Help: Non-Singleton Prediction Set}}\label{algo1:HelpTrigger1}
\State  \textcolor{black}{Ask human operator to choose decision from $\mathcal{C}(\ell(t+k))$\;}  \label{algo1:helphumanstart}
\If{\textcolor{black}{Human operator halts operation}} \Comment{\textcolor{black}{No correct action in $\ccalC({\ell}(t+k))$}} \label{algo1:humanhalt}
\State \textcolor{black}{Update  $\Sigma_{\text{unc}}$: $\Sigma_{\text{unc}}(t+k)\leftarrow\Sigma_{\text{unc}}(t+k)\cup\{\pi_{\text{next}}\}$ (if $\pi_{\text{next}}$ is not already in $\Sigma_{\text{unc}}(t+k)$)} \label{algo1:addSunc}
\State \textcolor{black}{Request new NL-based sub-task ($\pi_{\text{next}}$, $\Sigma_{\text{unsafe}}$) from LTL planner}  \label{algo1:newdfa1} 
\If{$\pi_{\text{next}} \neq \varnothing$}\label{algo1:helpdfa}
\State $t\gets t+k$, and go to line 5\;\label{algo1:helpdfa2}
\Else 
\State \textcolor{black}{Report mission failure (termination)} \label{algo1:report_fail0}
\EndIf
\Else
\State \textcolor{black}{Obtain $s(t+k)\in\mathcal{C}(\ell(t+k))$ from human operator\;}  \label{algo1:helphuman1}
\EndIf
\Else
\State  Pick (unique) decision $s(t+k)\in\mathcal{C}(\ell(t+k))$\label{algo1:pickAction}
\EndIf
\If {$\Omega$ prevents execution of $s(t+k)$} \Comment{\textcolor{black}{Ask-for-Help: Physical Environment}}\label{algo1:exec_fail}
\State \textcolor{black}{Update  $\Sigma_{\text{unc}}$: $\Sigma_{\text{unc}}(t+k)\leftarrow\Sigma_{\text{unc}}(t+k)\cup\{\pi_{\text{next}}\}$ (if $\pi_{\text{next}}$ is not already in $\Sigma_{\text{unc}}(t+k)$)} \label{algo1:addSunc2}
\State \textcolor{black}{Request new NL-based sub-task ($\pi_{\text{next}}$, $\Sigma_{\text{unsafe}}$) from LTL planner} \label{algo1:requestLTL2}
\If {$\pi_{\text{next}} \neq \varnothing$}
\State $t\gets t+k$, and go to line 5 \label{algo1:helpdfa_3}
\Else 
\State Report mission failure (termination) \label{algo1:report_fail}
\EndIf
\Else\label{algo1:elsec}
\State Execute decision $s(t+k)$ \label{algo1:run_command}
\State Observe environment update $p(t+k)$ \label{algo1:env_update}
\EndIf
\State Update $\ell(t+k+1)\gets \ell(t+k)+s(t+k)+p(t+k)$ \label{algo1:prompt_update}
\EndFor
\State \textcolor{black}{Remove all subtasks from $\Sigma_{\text{unc}}$ that were included because of LLM uncertainty (i.e., in line \ref{algo1:addSunc}).} \label{algo1:removeSunc}
\State Construct $\tau = [s(t),\dots,s(t+T-1)]$\label{algo1:s}
\State Append $\tau$ to the plan $\tau_{\phi}$\label{algo1:tau}
\State Current time step: $t\gets t+T$
\State Request new \textcolor{black}{NL-based sub-task ($\pi_{\text{next}}$, $\Sigma_{\text{unsafe}}$) from LTL planner} and check for mission accomplishment\label{algo1:missionProg}
\EndWhile
\end{algorithmic}
\end{algorithm}
\normalsize

In this section, we propose HERACLEs, a hierarchical planner 
to address the considered LTL-NL planning problem.
In Section \ref{sec:overview}, we present an overview of HERACLEs; see Alg. \ref{alg1:cp_ltl_planner}. A detailed description of its components is provided in Sections \ref{sec:LTLplan}-\ref{sec:labeling}. 

\vspace{-0.3cm}
\subsection{Overview of HERACLEs}\label{sec:overview}
HERACLEs takes as input an LTL-NL mission defined over a partially known semantic environment. This LTL-NL formula is processed online by an existing symbolic temporal logic planner that, given the current mission status, determines the next language-based sub-task the robot should accomplish to make mission progress. This symbolic planner is presented in Section \ref{sec:LTLplan} and is adopted from \cite{kantaros2020reactive,vasilopoulos2021reactive}; we note that any other temporal logic planner can be used. This language-based sub-task serves as an input to a pre-trained LLM that is responsible for generating a feasible plan of the form \eqref{eq:tau}; see Section \ref{sec:LLMplan}. \textcolor{black}{The LLM planner communicates back to the LTL planner with one of the following messages: (i) The LLM is `uncertain' about what the correct plan is [lines \ref{algo1:pred_set}-\ref{algo1:HelpTrigger1}, Alg. \ref{alg1:cp_ltl_planner}]. 
\textcolor{black}{In this case, help is requested from the human operator, who either provides the correct decision or chooses to halt the operation [lines \ref{algo1:helphumanstart}–\ref{algo1:helphuman1}, Alg.\ref{alg1:cp_ltl_planner}]. If the operator decides to halt the operation, the LTL planner attempts to generate an alternative sub-task to continue the mission [lines \ref{algo1:addSunc}–\ref{algo1:helpdfa2}, Alg.\ref{alg1:cp_ltl_planner}]. If no such sub-task exists, the mission terminates unsuccessfully (e.g., the LTL-NL mission may be infeasible) [line \ref{algo1:report_fail0}, Alg.~\ref{alg1:cp_ltl_planner}].
}
%
%
(ii) A plan has been designed but the (initially unknown) geometric structure of the environment ends up preventing the robot from fully executing it [line \ref{algo1:exec_fail}, Alg. \ref{alg1:cp_ltl_planner}]. 
In this case, the LTL planner generates an alternative sub-task to proceed, and the above process repeats [lines \ref{algo1:addSunc2}-\ref{algo1:helpdfa_3}, Alg. \ref{alg1:cp_ltl_planner}]. If such a sub-task does not exist, the mission terminates unsuccessfully [line \ref{algo1:report_fail}, Alg. \ref{alg1:cp_ltl_planner}]. 
(iii) A plan has been designed and executed, accomplishing the assigned sub-task [lines \ref{algo1:elsec}-\ref{algo1:env_update}, Alg. \ref{alg1:cp_ltl_planner}]. In this case, the LTL planner generates a new sub-task if the overall mission has not been accomplished yet, and the above process repeats [line \ref{algo1:missionProg}, Alg. \ref{alg1:cp_ltl_planner}].  
A detailed description of when the LLM requests assistance (cases (i)-(ii)) from the symbolic planner is provided in Section \ref{sec:seek4help}. Conformal prediction, outlined in Section \ref{sec:labeling}, is employed for uncertainty quantification in case (i).}

\vspace{-0.3cm}
\subsection{Symbolic Task Planner: What NL-based Sub-Task to Accomplish Next?
}\label{sec:LTLplan}
\vspace{-0.1cm}

\begin{wrapfigure}{r}{5cm}
  \centering \vspace{-4mm}
  \includegraphics[width=0.8\linewidth]{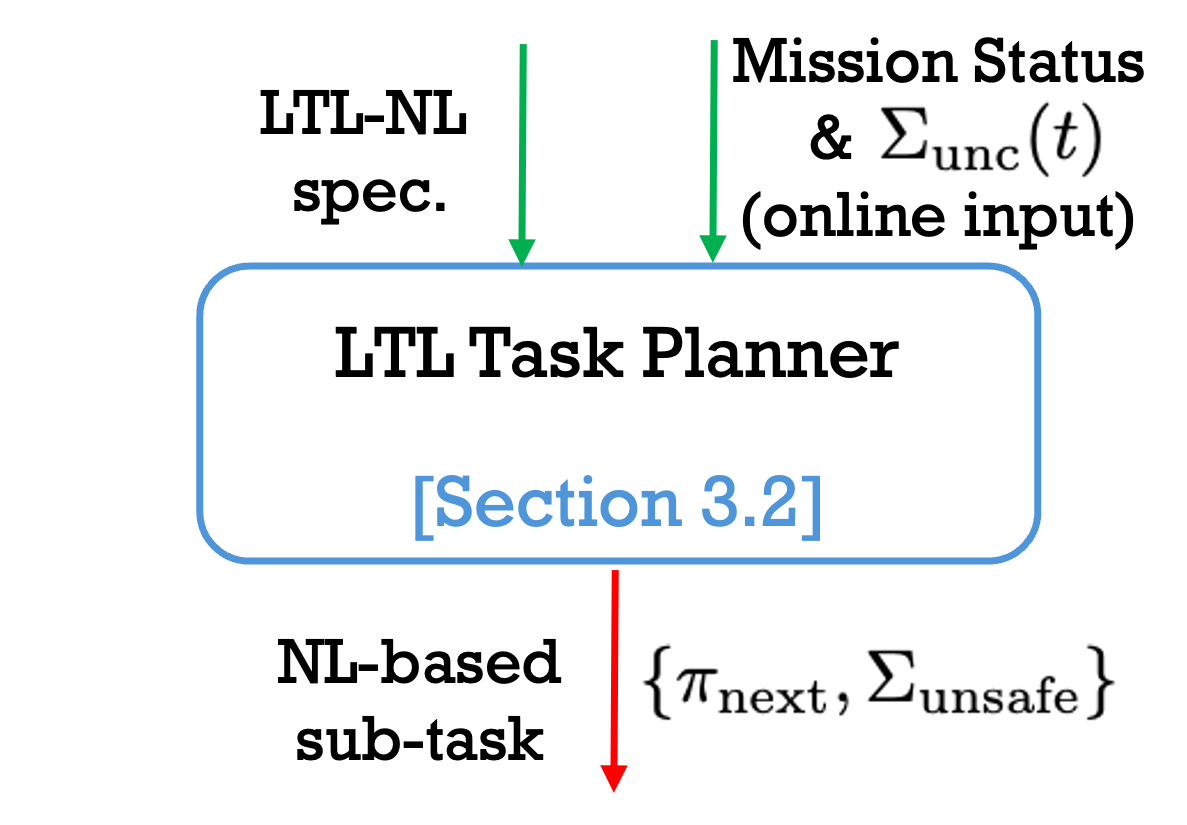} \vspace{-3mm}
  \caption{\textcolor{black}{Input-output structure of the symbolic planner. The output serves as an input to the neural planner (Fig. \ref{fig:main_body}).}
  }\vspace{-5mm} \label{fig:ltldecomp}
\end{wrapfigure}

In this section, we provide an overview of a symbolic/LTL task planner, presented in \cite{kantaros2020reactive,vasilopoulos2021reactive}, that we employ to determine what language sub-task the robot should accomplish next to make progress towards accomplishing the LTL-NL task $\phi$. \textcolor{black}{Its input-output structure is illustrated in Fig. \ref{fig:ltldecomp}; see also Ex. \ref{ex:me1}.} \textcolor{black}{We emphasize that any other temporal logic planner can be used as long as it is complete, i.e., if there exists a sub-task to make progress, the employed planner will compute it. 
} 

Consider \textcolor{black}{a mission scenario drawn from $\ccalD$ associated with} an LTL-NL specification $\phi$. Initially, we translate $\phi$  into a Deterministic Finite state Automaton (DFA) \cite{baier2008principles}; \textcolor{black}{this step is performed offline, before robot deployment}. \textcolor{black}{The DFA can be viewed as a directed graph,} with a set of nodes/states and \textcolor{black}{edges/transitions} 
between states. It has an initial node representing the mission's commencement, a set of final nodes denoting mission completion, and intermediate nodes representing various stages of the mission. Transitions between nodes occur when specific atomic predicates are satisfied. Mission accomplishment is achieved upon reaching a final \textcolor{black}{node while starting from the initial node}.

The \textcolor{black}{symbolic} task planner receives three inputs: (i) an LTL-NL task $\phi$; (ii)  the current mission status at time $t$; and (iii) a set $\Sigma_{\text{unc}}(t)$ of NL-based APs. The input (ii) is represented by the DFA state that has been reached, starting from the initial one, given the sequence of actions that the robot has applied up to the current time step $t$. 
\textcolor{black}{Initially, $\Sigma_{\text{unc}}(0)=\emptyset$ [line \ref{algo1:dfa}, Alg. \ref{alg1:cp_ltl_planner}]. As the robot navigates, $\Sigma_{\text{unc}}$ collects each NL-based AP for which either the LLM is uncertain about how to design a feasible plan, or whose satisfaction is physically impossible due to the (partially unknown) geometry of the environment.}
%
%
The details of its update are discussed in Section \ref{sec:seek4help}.
Given these inputs, the \textcolor{black}{symbolic} task planner generates: (i) an NL-based AP, denoted by $\pi_{\text{next}}$; and (ii) a set of NL-based APs collected in a set $\Sigma_{\text{unsafe}}$ [line \ref{algo1:dfa}, Alg. \ref{alg1:cp_ltl_planner}]. These sets impose two conditions that must be met to make mission progress (i.e., to move closer to a final DFA node). Specifically, the robot should design a plan $\tau$, defined as in 
\eqref{eq:tau}, to (a) accomplish the NL-based sub-task captured in $\pi_{\text{next}}$ while, (b) in the meantime, not satisfying any of the APs collected in $\Sigma_{\text{unsafe}}$. Note that the set $\Sigma_{\text{unsafe}}$ depends on the current mission status and, therefore, it does not remain fixed throughout the mission. These two requirements act as inputs to an LLM-based planner that is responsible for designing a plan $\tau$ that satisfies (a)-(b); see Section \ref{sec:LLMplan}. 

\begin{wrapfigure}{r}{10cm}
  \centering \vspace{-4mm}
  \includegraphics[width=1\linewidth]{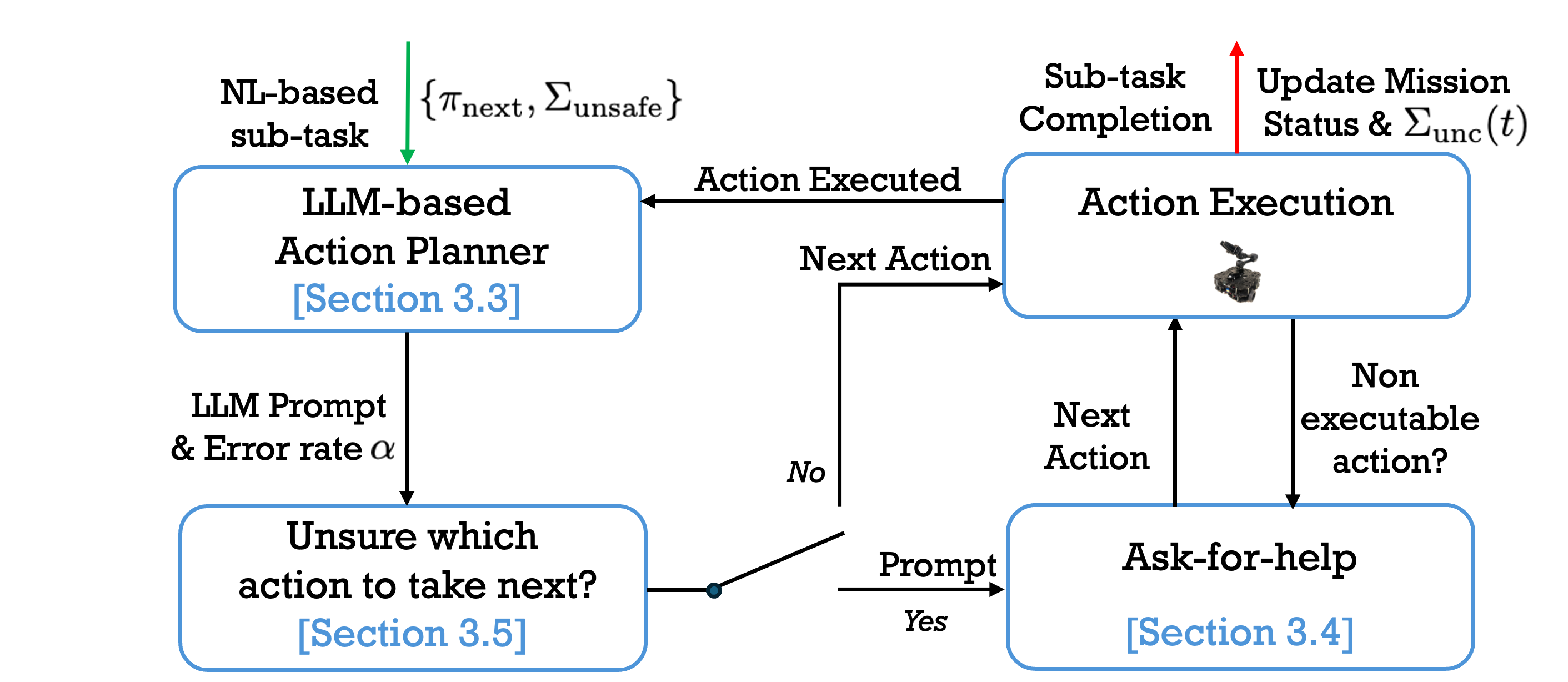}\\ \vspace{-3mm}
  \caption{\textcolor{black}{Overview of the neural planner. Given an NL-encoded sub-task from the symbolic planner, the neural planner generates robot actions and requests help—from either the symbolic planner or the user (see Fig.~\ref{fig:help_module})—in cases of high uncertainty or infeasible sub-tasks. Once the sub-task is completed, the mission status and the set $\Sigma_{\text{unc}}$ are updated back to the symbolic planner.}
  }\vspace{-5mm} \label{fig:main_body}
\end{wrapfigure}

\textcolor{black}{Notice that the temporal logic planner discussed above is myopic in the sense that it computes the next NL-based sub-task $\pi_{\text{next}}$ for the robot, instead of a sequence of sub-tasks that need to be accomplished to satisfy the overall mission. A necessary condition required to ensure completeness of that symbolic planner is that the APs are independent of each other, which means that the satisfaction of one AP does not prohibit the satisfaction of another AP. This assumption can be relaxed by employing non-myopic symbolic planners, which generate a sequence of sub-tasks (rather than only the next sub-task)  that the robot should accomplish to complete the mission \cite{kantaros2020stylus, shoukry2017linear}. The benefit of using the employed myopic symbolic planner, as opposed to a non-myopic one, lies in its computational efficiency, as demonstrated in \cite{kantaros2020reactive, vasilopoulos2021reactive}.}
%
%

\vspace{-0.1cm}
\begin{ex}[Symbolic Planner]
\textcolor{black}{Consider the LTL-NL formula $\phi$ discussed in Example~\ref{ex:me}. At time $t = 0$, the symbolic planner generates the sub-task specified by $\pi_{\text{next}} = \pi_1$ and sets $\Sigma_{\text{unsafe}} = \{\pi_4\}$. 
}
%
%
\label{ex:me1}
\end{ex}

\vspace{-0.6cm}
\subsection{LLM-based Action Planner: How to Accomplish the Assigned NL-based Sub-Task?
}\label{sec:LLMplan}

%
\textcolor{black}{In this section, we present our neural planner, that employs pre-trained LLMs, to synthesize a finite horizon plan $\tau$ satisfying conditions (a)-(b) \textcolor{black}{(see Fig. \ref{fig:main_body}).}} 
%
%
%
Hereafter, we simply refer to the requirements (a)-(b) as the (NL-based) sub-task that the robot should accomplish next.  Inspired by \cite{ahn2022can,ren2023robots}, we convert this planning problem into a sequence of $T>0$ multiple-choice question-answering (MCQA) problems for the LLM where $T$ is a hyperparameter.\footnote{Essentially, $\tau$ and $T$ refer to $\tau_n$ and $T_n$ defined in Section \ref{sec:problem}. For ease of presentation, we drop the dependence on $n$.}  The `question' refers to the sub-task along with any actions that the robot has taken up to time $t+k$, $k\in\{0,\dots,T-1\}$ to accomplish it, where $t$ stands for time step when the sub-task was announced by the symbolic planner. We denote by $\ell(t+k)$ the (textual) description of the `question'. The `choices' refer to the decisions $s(t+k)$ that the robot can make.  Given $\ell(t+k)$, the LLM will select $s(t+k)$. This process occurs sequentially for $k\in\{0,\dots,T-1\}$ giving rise to a plan $\tau$. Next, we discuss how $\ell(t+k)$ is structured and be used to compute $s(t+k)$.

\begin{figure}[t] 
\centering
\includegraphics[width=0.7\linewidth]{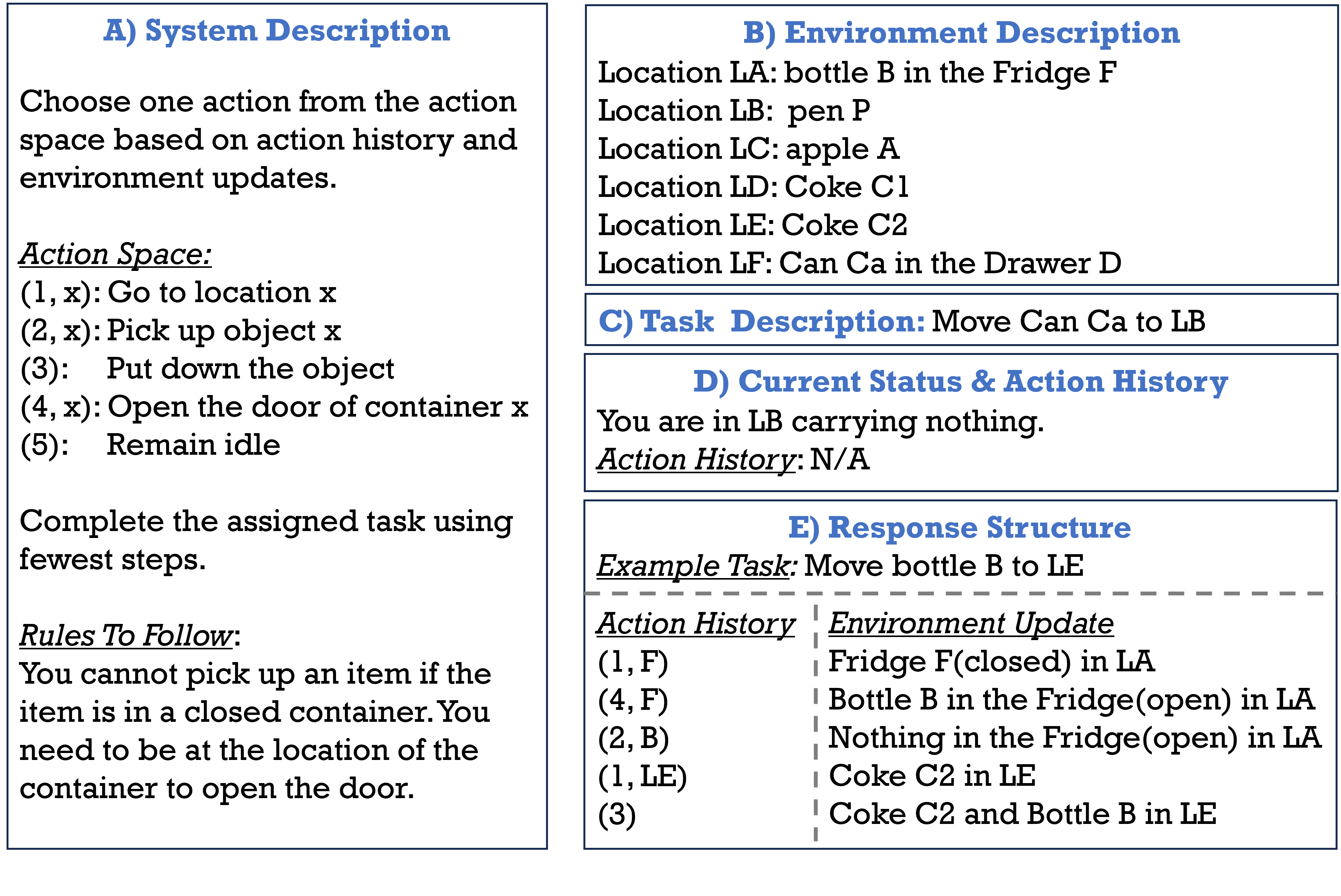}  \vspace{-0.8cm}
\caption{Example of a prompt constructed in our simulations in Section \ref{sec:experiments} using GPT 3.5. This prompt refers to $t=1$ when the history of actions is empty. 
} 
\label{fig:prompt}\vspace{-0.6cm}
\end{figure}

\textbf{Prompt Construction:} Here we discuss how we structure the context text (`prompt') that will be given as an input to the LLM; see also Fig. \ref{fig:prompt}.  The prompt in this work consists of the following parts. (A) \textit{System description} that defines the action space determining all possible actions $a\in\ccalA$ that the robot can apply as well as the objective of the LLM. 
(B) \textit{Environment description} that describes the \textcolor{black}{initial} locations $\bbx_e$ of each semantic object $e$ of interest;
(C) \textit{Task description} that includes the language-based task $\pi_{\text{next}}$ (condition (a)) as well as language-based constraints (if any), modeled by $\Sigma_{\text{unsafe}}$ that the robot should respect until $\pi_{\text{next}}$ is satisfied (condition (b)); 
(D) \textit{History of actions \& current environment status} that includes the sequence of actions, generated by the LLM, that the robot has executed so far towards accomplishing the assigned sub-task. It also includes the current locations of semantic objects that the robot may have manipulated/moved so far; 
(E) \textit{Response structure} describing the desired structure of the LLM output for an example task.

\textbf{Plan Design \& Execution:} 
At iteration $k=0$, part (D) does not include any textual information as a new sub-task has just been announced [line \ref{algo1:prompt_init}, Alg. \ref{alg1:cp_ltl_planner}]. 
Given $\ell(t+k)$, the LLM is asked to make a decision $s(t+k)$ among all available ones included in part (A). 
The LLM selects $s(t+k)$ as follows. Given any $s\in\ccalS$, LLMs can provide a logit/confidence score $g(s|\ell(t+k))$; the higher the score, the more likely the decision $s$ is a valid next step to address the language-instruction provided in $\ell(t+k)$. To get $g(s|\ell(t+k))$, we query the LLM over all potential decisions $s\in\ccalS$ \cite{ahn2022can}.

\begin{wrapfigure}{r}{6cm}
  \centering \vspace{-4mm}
  \includegraphics[width=1\linewidth]{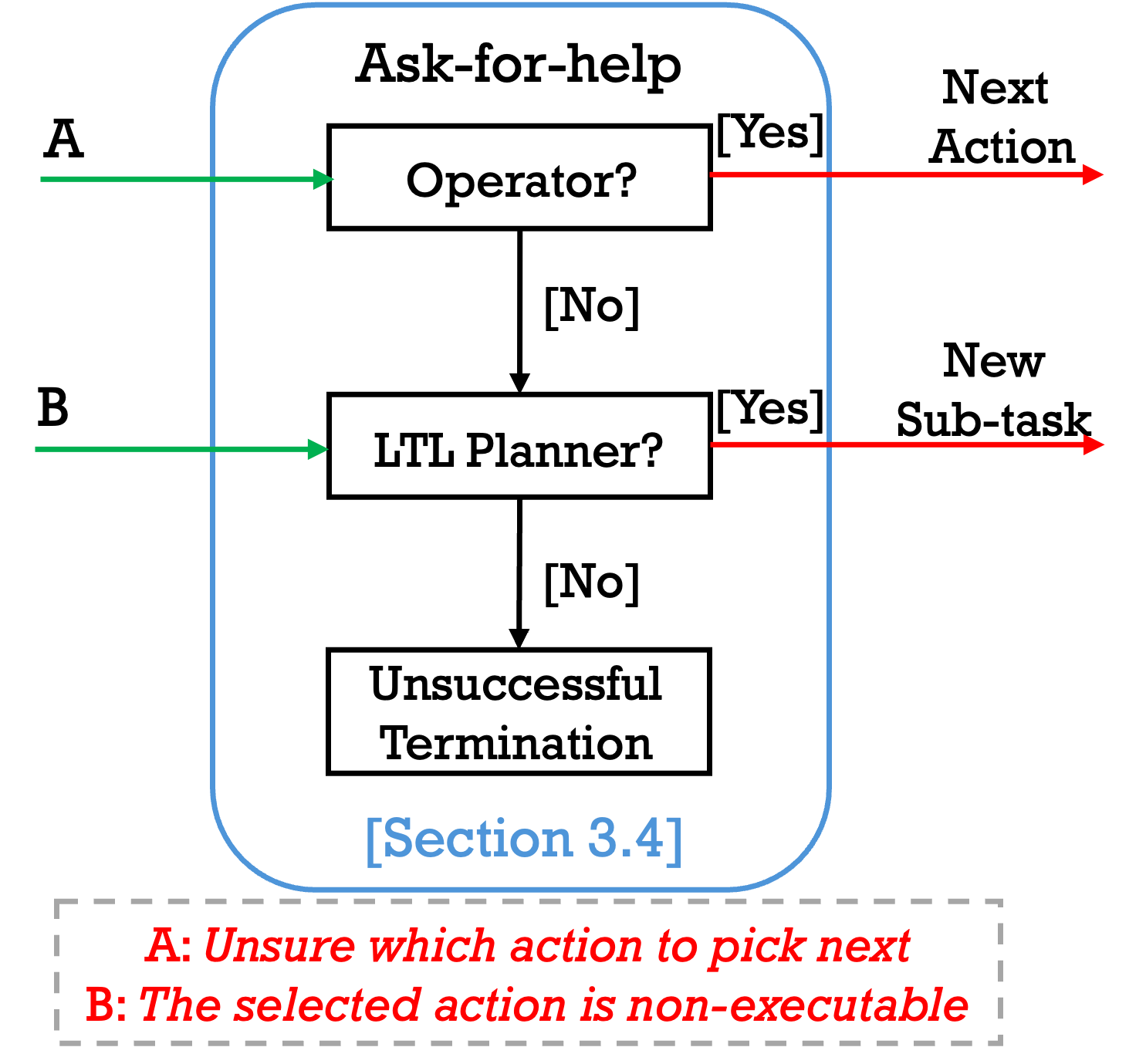}\\ \vspace{-4mm}
  \caption{\textcolor{black}{Graphical overview of the help module. The type of assistance provided depends on the condition that triggered the help mode (case A or B). A "request new sub-task" message is sent to the symbolic planner (not shown in Fig. \ref{fig:ltldecomp}), while the "next action" is forwarded to the action execution module in Fig. \ref{fig:main_body}.}
  }\vspace{-7mm} \label{fig:help_module}
\end{wrapfigure}

Using these scores, a possible approach to select $s(t+k)$ is by simply choosing the decision with the highest score, i.e., $s(t+k)=\arg\max_{s\in\mathcal{S}} g(s|\ell(t+k))$. However, these scores do not represent calibrated confidence. 
Thus, we calibrate these confidence scores and let the LLM make decisions only when it is certain enough \cite{ren2023robots}. We formalize this by constructing a set of actions (called, hereafter, \textit{prediction set}), denoted by $\ccalC(\ell(t+k))$, that contain the ground truth action with user-specified confidence determined by the user-specified mission success rate $1-\alpha$ defined in Section \ref{sec:problem} [line \ref{algo1:pred_set}, Alg. \ref{alg1:cp_ltl_planner}]. 
Hereafter, we assume that such prediction sets are provided; we defer their construction, using conformal prediction, to Section \ref{sec:labeling} (see \eqref{eq:pred3}). Given $\ccalC(\ell(t+k))$, we select the decision $s(t+k)$ as follows. If $|\ccalC(\ell(t+k))|=1$, then we select the decision included in $\ccalC(\ell(t+k))$ as it contributes to mission progress with high confidence \textcolor{black}{[line \ref{algo1:pickAction}, Alg. \ref{alg1:cp_ltl_planner}]}. \textcolor{black}{By construction of the prediction sets, this action coincides with $s(t+k)=\arg\max_{s\in\mathcal{S}} g(s|\ell(t+k))$; see Section \ref{sec:labeling}.} As soon as it is selected, the robot physically executes it using its library of low-level controllers \textcolor{black}{[line \ref{algo1:run_command}, Alg. \ref{alg1:cp_ltl_planner}]}. If $|\ccalC(\ell(t+k))|>1$ \textcolor{black}{[line \ref{algo1:HelpTrigger1}-\ref{algo1:helphuman1}, Alg.\ref{alg1:cp_ltl_planner}]} \textcolor{black}{or 
the robot cannot physically execute $s(t+k)$}, the robot seeks assistance to select $s(t+k)$ \textcolor{black}{[line \ref{algo1:exec_fail}-\ref{algo1:report_fail}, Alg.\ref{alg1:cp_ltl_planner}]. The process of asking for assistance is described in more detail in Section \ref{sec:seek4help}} \textcolor{black}{and illustrated in Fig. \ref{fig:help_module}.} 

Once a decision $s(t+k)$ is made and executed, the current time step is updated to $t+k+1$. Then, the prompt $\ell(t+k+1)$ is constructed that will be used to select $s(t+k+1)$. Parts (A)-(C) and (E) in the prompts $\ell(t+k)$ and $\ell(t+k+1)$ are the same. Part (D) in $\ell(t+k+1)$ is augmented by recording the decision $s(t+k)$ as well as incorporating perceptual feedback about the status of containers (if any) that may contain any of the semantic objects. We automatically convert this perceptual feedback into text denoted by $p(t+k)$ (see Fig. \ref{fig:prompt}) [line \ref{algo1:env_update}, Alg. \ref{alg1:cp_ltl_planner}]. 
With slight abuse of notation, we denote this prompt update by
$\ell(t+k+1)=\ell(t+k)+s(t+k)+p(t+k)$
%
where summation means text concatenation [line \ref{algo1:prompt_update}, Alg.\ref{alg1:cp_ltl_planner}]. This process is repeated for all $k\in\{0,\dots,T-1\}$ to sequentially select $s(t+k)$.
%
%
%
This process generates a plan $\tau$ 
%
%
of length $T$ as in \eqref{eq:tau} [line \ref{algo1:s}, Alg. \ref{alg1:cp_ltl_planner}].
At time $t+T$, if the LTL-NL mission has not been accomplished yet, the LTL task planner generates a new sub-task and the above process repeats [line \ref{algo1:missionProg}, Alg. \ref{alg1:cp_ltl_planner}].  
The concatenation of all plans $\tau$ for the sub-tasks generated by the LTL planner results in the plan $\tau_{\phi}$ in \eqref{eq:tauphi} [line \ref{algo1:tau}, Alg. \ref{alg1:cp_ltl_planner}]. 

\vspace{-0.4cm}
\subsection{When to Seek Assistance?} \label{sec:seek4help}
%
Assume that there exists at least one $k\in\{0,\dots,T-1\}$ so that $|\mathcal{C}(\ell(t+k)))|> 1$. In this case, the robot asks for help in order to proceed [lines \ref{algo1:HelpTrigger1}-\ref{algo1:helphuman1}, Alg. \ref{alg1:cp_ltl_planner}]. \textcolor{black}{The help process occurs as follows; see Fig. \ref{fig:help_module} and Ex. \ref{ex:me2}. 
Initially, the robot requests help from the human operator [line \ref{algo1:helphumanstart}, Alg. \ref{alg1:cp_ltl_planner}]. Specifically, the robot presents the current prompt $\ell(t+k)$  to the user along with the prediction set $\mathcal{C}(\ell(t+k)))$. The user selects the correct decision $s(t+k)$ if it exists in the prediction set [line \ref{algo1:helphuman1}, Alg. \ref{alg1:cp_ltl_planner}]. If no correct decision is available, the user halts the operation, and the robot will then request from the LTL planner an alternative sub-task to make mission progress \textcolor{black}{[lines \ref{algo1:humanhalt}-\ref{algo1:report_fail0}, Alg. \ref{alg1:cp_ltl_planner}]}. To make this help-request, the set $\Sigma_{\text{unc}}$, defined in Section \ref{sec:LTLplan}, is updated as $\Sigma_{\text{unc}}=\Sigma_{\text{unc}}\cup\{\pi_{\text{next}}\}$ \textcolor{black}{[line \ref{algo1:addSunc}, Alg. \ref{alg1:cp_ltl_planner}]}. The resulting set $\Sigma_{\text{unc}}$, along with the current mission status, is sent to the LTL task planner, which then generates a new sub-task [line \ref{algo1:newdfa1}, Alg. \ref{alg1:cp_ltl_planner}]. If such a sub-task exists, the process discussed in Section \ref{sec:LLMplan} repeats [line \ref{algo1:helpdfa2}, Alg. \ref{alg1:cp_ltl_planner}].
If there are no alternative sub-tasks to proceed, the LTL planner fails to assist. 
Help from the LTL planner is also requested if the robot cannot physically execute the selected decision $s(t+k)$. This may occur when the robot detects that the initially unknown geometry of the environment prevents it from reaching the desired destinations or objects, as in \cite{kantaros2020reactive,vasilopoulos2021reactive} [lines \ref{algo1:exec_fail}–\ref{algo1:report_fail}, Alg. \ref{alg1:cp_ltl_planner}]. In such cases, $\pi_{\text{next}}$ is first added to $\Sigma_{\text{unc}}$ [line \ref{algo1:addSunc2}, Alg. \ref{alg1:cp_ltl_planner}], and then the LTL planner is asked to generate an alternative sub-task. If no alternative sub-task can be generated, the mission terminates unsuccessfully [line \ref{algo1:report_fail}, Alg. \ref{alg1:cp_ltl_planner}]. At time step $t+T$, i.e., upon completion of the current sub-task, all sub-tasks previously added to $\Sigma_{\text{unc}}$ due to non-singleton prediction sets not containing the ground truth decision are removed from that set [line~\ref{algo1:removeSunc}, Alg.~\ref{alg1:cp_ltl_planner}]. Instead, sub-tasks added to $\Sigma_{\text{unc}}$ because their execution was physically blocked by the environment remain in the set permanently. This ensures that the symbolic planner will no longer recommend sub-tasks that are physically impossible to complete due to the environment's geometric structure.
}
\begin{ex}[Neural Planner \& Help Mode]
\textcolor{black}{Consider the LTL-NL formula of Ex.~\ref{ex:me} and the sub-task, modeled by $\pi_{\text{next}} = \pi_1$ and $\Sigma_{\text{unsafe}} = \{\pi_4\}$ as generated by the symbolic planner in Ex. \ref{ex:me1}. The neural planner initially generates the action `go to Cone 1' to complete $\pi_1$. However, once $\pi_1$ is deemed infeasible—due to cone 1 being inaccessible (see Fig.~\ref{fig:env})—we request an alternative sub-task from the symbolic planner.
In response, the symbolic planner generates $\pi_{\text{next}} = \pi_3$, while keeping the unsafe set unchanged, i.e., $\Sigma_{\text{unsafe}} = \{\pi_4\}$. The set $\Sigma_{\text{unc}}$, which was initially empty, is updated to $\Sigma_{\text{unc}}(t) = \{\pi_1\}$.  
}
\label{ex:me2}
\end{ex}

\vspace{-0.2cm}
\subsection{Constructing Prediction Sets 
}\label{sec:labeling}
\vspace{-0.1cm}

\begin{algorithm}[t]
\footnotesize
\caption{Computation of Causal Prediction Sets $\mathcal{C}({\ell}_{\text{test}}(t))$ 
}\label{alg2:alg2}
\begin{algorithmic}[1]
\State \textcolor{black}{\textbf{Inputs}: Calibration set $\ccalM$;  Current prompt ${\ell}_{\text{test}}(t)$; Threshold $\alpha$}
\State \textcolor{black}{\textbf{Output}: Causal Prediction Set $\mathcal{C}({\ell}_{\text{test}}(t))$}
\If{$t=1$}
\State \textcolor{black}{Compute $\bar{r}_i$ for all calibration sequences $i$ (see \eqref{eq:barri})}\label{algo2:compute_r_i}
\State \textcolor{black}{Compute the $\frac{(M+1)(1-\alpha)}{M}$ quantile $\bar{q}$ of $\{\bar{r}_i\}_{i=1}^{M}$} \label{algo2:compute_quantile}
\EndIf
\State \textcolor{black}{
Compute the LLM confidence $g(s|{\ell}_{\text{test}}(t))$, for each $s\in\ccalS$ \label{algo2:compute_confidence}}
\State Return the prediction set $\mathcal{C}({\ell}_{\text{test}}(t))=\{s~|~g(s|{\ell}_{\text{test}}(t))>1-\bar{q}\}$ (see \eqref{eq:causalPred}). \label{algo2:get_pred_set}
\end{algorithmic}
\end{algorithm}
\normalsize

In this section, we discuss how the prediction sets $\ccalC(\ell(t))$, introduced in Section \ref{sec:LLMplan}, are constructed. To construct them, we employ conformal prediction (CP), a statistical tool for uncertainty quantification in black box models \cite{angelopoulos2023conformal}. To illustrate the challenges in their construction, we consider two cases: (i) single-step plans, 
and (ii) multi-step plans. Our analysis builds upon \cite{ren2023robots,kumar2023conformal}. \textcolor{black}{In what follows, for simplicity, we assume that there exists a unique correct plan for each scenario drawn from $\ccalD$.} \textcolor{black}{This assumption is relaxed in Appendix \ref{sec:RelaxI}.} 

\textbf{Single-step Plans:} We begin by considering LTL-NL formulas $\phi$ that can be satisfied by plans $\tau_{\phi}$ of horizon $H=1$ (see Section \ref{sec:problem}); later we generalize the results for $H\geq1$. This also means that synthesizing  $\tau_{\phi}$ requires the LLM to make a single decision $s$. 
%
%
First, we sample $M$ independent scenarios from $\ccalD$. We refer to these scenarios as calibration scenarios. For each calibration scenario $i\in\{1,\dots,M\}$, we construct its equivalent prompt $\ell_{\text{calib}}^i$. For each prompt, we (manually) compute the ground truth plan $\tau_{\text{calib}}^{i}=s_{\text{calib}}^{i}(1)$ accomplishing this task. 
%
%
Hereafter, we drop the dependence on the robot decisions and prompts on the time step, since we consider single-step plans. This way we construct a calibration dataset $\ccalM=\{(\ell_{\text{calib}}^i,\tau_{\text{calib}}^{i})\}_{i=1}^M$.

Consider a new scenario drawn from $\ccalD$, called validation/test scenario. We convert this scenario into its equivalent prompt $\ell_{\text{test}}$. Since the calibration and the validation scenario are i.i.d., 
%
CP can generate a prediction set $\ccalC(\ell_{\text{test}})$ of decisions $s$ containing the correct one $s_{\text{test}}$ with probability greater than $1-\alpha$, i.e., 
\begin{equation}\label{eq:CP1}
P(s_{\text{test}}\in \mathcal{C}(\ell_{\text{test}})) \geq 1-\alpha,
\end{equation}
where $\alpha\in (0,1)$ is user-specified.
To generate $\mathcal{C}(\ell_{\text{test}})$, CP first uses the LLM’s  confidence score $g$ (see Section \ref{sec:LLMplan}) to compute the set of
non-conformity (NC) scores $\{r_i=1-g(s_{\text{calib}}^i~|~\ell_{\text{calib}}^i)\}_{i=1}^M$ over the calibration set. 
%
The higher the \textcolor{black}{NC} score $r_i$ is, the worse the performance of the LLM is at the $i$-th calibration point.
Then CP performs calibration by computing the $\frac{(M+1)(1-\alpha)}{M}$ 
%
empirical quantile of $r_1,\dots,r_M$ denoted by $q$. Then, it generates the  prediction set 
%
\begin{equation}\label{eq:pred1}
\mathcal{C}(\ell_{\text{test}})=\{s\in \ccalS~|~g(s|\ell_{\text{test}})>1-q\},
\end{equation}
that includes all decisions that the predictor is at least $1-q$ confident in. The generated prediction set ensures the $1-\alpha$ \textit{marginal} coverage guarantee in \eqref{eq:CP1} holds. 
%
This coverage guarantee is marginal, defined over the randomness of the calibration data and validation scenario.
%
%
By construction of the prediction sets the decision $s=\arg\max_{s\in\mathcal{S}} g(s|\ell_{\text{test}})$ belongs to $\mathcal{C}(\ell_{\text{test}})$. 

\textbf{Multi-step Plans:} Next, we generalize the above result to the case where satisfaction of $\phi$ requires plans $\tau_{\phi}$ with $H\geq 1$ decisions selected from $\ccalS$; \textcolor{black}{see Alg. \ref{alg2:alg2}.} Here we cannot apply directly \eqref{eq:CP1}-\eqref{eq:pred1} to compute individual sets $\ccalC(\ell_{\text{test}}(t))$ for the robot, as this violates the i.i.d. assumption required to apply CP. The challenge in this case is that the prompts $\{(\ell_{\text{test}}(t)\}_{t=1}^H$ are not independent of each other which violates the i.i.d. assumption required to apply CP. In fact these prompts depend on past robot decisions as well as on the LTL-NL tasks $\phi_{\text{test}}$. To address this challenge, inspired by \cite{ren2023robots}, we (i) lift the data to sequences, and (ii) perform calibration at the sequence level using a carefully designed NC score function. 

First, we construct a calibration dataset as follows. We generate $M\geq1$ scenarios $\xi_i$ from $\ccalD$. The LTL-NL formula $\phi_i$ of each scenario is broken into a sequence of $H_i\geq 1$ prompts, defined as:\footnote{The distribution $\ccalD$ over scenarios induces a distribution over data sequences \eqref{eq:seqProm} \cite{ren2023robots}. These data sequences are equivalent representations of the sampled scenarios augmented with the ground truth decisions.}
\begin{equation}\label{eq:seqProm}
\bar{\ell}_{\text{calib}}^i=[{\ell}_{\text{calib}}^i(1),\dots, {\ell}_{\text{calib}}^i(H_i)],
\end{equation}
where each prompt in the sequence $\bar{\ell}_{\text{calib}}^i$ contains a history of ground truth decisions made so far. We define the corresponding sequence of ground truth decisions as:
\begin{equation}
\tau_{\phi,\text{calib}}^i=s_{\text{calib}}^i(1),\dots,s_{\text{calib}}^i(H_i),
\end{equation}
This gives rise to the calibration set  $\ccalM=\{(\bar{\ell}_{\text{calib}}^i,\tau_{\phi,\text{calib}}^i)\}_{i=1}^M$. 
%
%
%
%
Next, we use the lowest score $g$ over the time-steps $1,\dots, H_i$ as the score for each sequence $i$ in calibration set, i.e., 
%
\begin{equation}\label{eq:min_seqpred}
\bar{g}(\tau_{\phi,\text{calib}}^i~|~\bar{\ell}_\text{calib}^i)=\min_{t\in\{1,\dots, H_i\}}g(\tau_{\phi,\text{calib}}^i(t)~|~{\ell}_\text{calib}^i(t)).
\end{equation}
Thus, the NC score of each sequence $i$ is \textcolor{black}{[line \ref{algo2:compute_r_i}, Alg. \ref{alg2:alg2}]}
\begin{equation}\label{eq:barri}
\bar{r}_i=1-\bar{g}(\tau_{\phi,\text{calib}}^i|\bar{\ell}_\text{calib}^i).
\end{equation}

Consider a new scenario $\xi_{\text{test}}$ associated with a task $\phi_{\text{test}}$ with horizon $H_{\text{test}}$. This scenario corresponds to a sequence of prompts
\begin{equation}
    \bar{\ell}_{\text{test}}={\ell}_{\text{test}}(1),\dots,{\ell}_{\text{test}}(k),\dots,{\ell}_{\text{test}}(H_{\text{test}}).
\end{equation}
CP can generate a prediction set $\bar{\ccalC}(\bar{\ell}_{\text{test}})$ of plans $\tau_{\phi}$ containing the correct one $\tau_{\phi,\text{test}}$  with high probability i.e., 
\begin{equation}\label{eq:CP2}
P(\tau_{\phi,\text{test}}\in \bar{\mathcal{C}}(\bar{\ell}_{\text{test}})) \geq 1-\alpha,
\end{equation}
where the prediction set $\bar{\mathcal{C}}(\bar{\ell}_{\text{test}})$ is defined as 
%
\begin{equation}\label{eq:pred3}
\bar{\mathcal{C}}(\bar{\ell}_{\text{test}})=\{\tau_{\phi}~|~\bar{g}(\tau_{\phi}|\bar{\ell}_{\text{test}})>1-\bar{q}\}, 
\end{equation}
where $\bar{q}$ is the $\frac{(M+1)(1-\alpha)}{M}$ 
empirical quantile of $\bar{r}_1,\dots,\bar{r}_M$ [line \ref{algo2:compute_quantile}, Alg. \ref{alg2:alg2}]. The generated prediction set ensures that the coverage guarantee in \eqref{eq:CP2} holds. By construction of the prediction sets, the plan $\tau_{\phi}$ generated by the LLM belongs to $\bar{\mathcal{C}}(\bar{\ell}_{\text{test}})$.

\textbf{Causal Construction of the Prediction Set:} Notice that $\bar{\mathcal{C}}(\bar{\ell}_{\text{test}})$ is constructed after the entire sequence $\bar{\ell}_{\text{test}}={\ell}_{\text{test}}(1),\dots,{\ell}_{\text{test}}(H_{\text{test}})$ is obtained. However, at every (test) time $t\in\{1,\dots,H_{\text{test}}\}$, the robot
%
observes only the prompt ${\ell}_{\text{test}}(t)$ and not the whole sequence. In what follows, we construct the prediction set in a causal manner using only the current and past information. Specifically, at every time step $t$, we construct the local prediction set as \textcolor{black}{[line \ref{algo2:compute_confidence}-\ref{algo2:get_pred_set}, Alg. \ref{alg2:alg2}]}  
%
%
\begin{equation}\label{eq:causalPred}
    \mathcal{C}({\ell}_{\text{test}}(t))=\{s~|~g(s|{\ell}_{\text{test}}(t))>1-\bar{q}\}.
\end{equation}
Then, the causal prediction set for $\bar{\ell}_{\text{test}}$ is defined as  
\begin{equation}\label{eq:causalCP}
    \mathcal{C}(\bar{\ell}_{\text{test}})=\mathcal{C}({\ell}_{\text{test}}(1))\times\mathcal{C}({\ell}_{\text{test}}(2))\times\dots\times\mathcal{C}({\ell}_{\text{test}}(H_{\text{test}})).
\end{equation}
In Section \ref{sec:thm}, we show that $\bar{\mathcal{C}}(\bar{\ell}_{\text{test}})=\mathcal{C}(\bar{\ell}_{\text{test}})$.

\vspace{-0.3cm}
\section{Probabilistic Task Satisfaction Guarantees} \label{sec:thm}

In this section, we show that given any (validation) scenario $\xi_{\text{test}}=\{\ccalA_{\text{test}}, \phi_{\text{test}}, H_{\text{test}}, \Omega_{\text{test}}\}$ drawn from \textcolor{black}a distribution $\ccalD$, the probability that HERACLEs will design a plan satisfying the LTL-NL formula is at least $1-\alpha$, where $\alpha$ is the coverage level used to construct the prediction sets.  \textcolor{black}{Recall from Section \ref{sec:labeling} that we assume that all mission scenarios drawn from $\ccalD$ admit a unique feasible plan. This means that only one sequence of sub-tasks can lead to mission completion, and there is a single robot plan that can successfully execute them. A discussion on how this assumption can be relaxed can be found in Appendix \ref{sec:RelaxI}.} To show this, we need first to show the following result. The proofs follow a similar logic as in \cite{ren2023robots}.

\begin{prop}\label{prop:eq_prop}
The prediction set $\bar{\ccalC}(\bar{\ell}_{\text{test}})$ defined in \eqref{eq:pred3} is the same as the on-the-fly constructed prediction set $\ccalC(\bar{\ell}_{\text{test}})$ defined in \eqref{eq:causalCP}, i.e.,  $\bar{\mathcal{C}}(\bar{\ell}_{\text{test}})=\mathcal{C}(\bar{\ell}_{\text{test}})$.
\end{prop}
\begin{proof}
It suffices to show that if $\tau_{\phi}\in\bar{\ccalC}(\bar{\ell}_{\text{test}})$ then $\tau_{\phi}\in{\ccalC}(\bar{\ell}_{\text{test}})$ and vice-versa. 
First, we show that if $\tau_{\phi}\in\bar{\ccalC}(\bar{\ell}_{\text{test}})$ then $\tau_{\phi}\in\ccalC(\bar{\ell}_{\text{test}})$. Since $\tau_{\phi}\in \bar{\mathcal{C}}(\bar{\ell}_{\text{test}})$, then we have that $$\min_{t\in\{1,\dots,H_{\text{test}}\}}g(\tau_{\phi}(t) | {\ell}_{\text{test}}(t) )>1-\bar{q},$$ due to \eqref{eq:min_seqpred}. This means that \textcolor{black}{the score} $g(\tau_{\phi}(t) | {\ell}_{\text{test}}(t) )>1-\bar{q}$, for all  $t\in \{1,\dots,H_{\text{test}}\}$. Thus, $\tau_{\phi}(t)\in\ccalC({\ell}_{\text{test}}(t))$, for all  $t\in \{1,\dots,H_{\text{test}}\}$. By definition of $\ccalC(\bar{\ell}_{\text{test}})$ in \eqref{eq:causalCP} such that $\tau_{\phi}\in\ccalC(\bar{\ell}_{\text{test}})$. These steps hold in the other direction too showing that if $\tau_{\phi}\in\ccalC(\bar{\ell}_{\text{test}})$ then $\tau_{\phi}\in\bar{\ccalC}(\bar{\ell}_{\text{test}})$. 
\end{proof}


\begin{theorem}\label{thm1}
Assume that \textcolor{black}{all scenarios drawn from $\ccalD$ have a unique feasible plan and that the prediction sets are constructed causally with coverage level $1-\alpha$.} 
%
\textcolor{black}{If (i) the LTL task planning algorithm is complete—that is, it can always generate an alternative sub-task, if it exists, whenever the current one is infeasible due to environmental structure, and it can generate a new sub-task as soon as the current one is completed—and (ii) Assumption~\ref{as:A1} holds,
then the completion rate over new test scenarios $\xi_{\text{test}}$ (and over the randomness of the calibration sets), drawn from $\ccalD$, is guaranteed to be at least $1 - \alpha$.}
\end{theorem}

\begin{proof}
%
%
If the symbolic planning algorithm is complete, then this means if there exists a solution it will find it. By solution, here we refer to a sequence of sub-tasks in the unknown environment that, if completed, the LTL mission will be satisfied. Since the scenarios $\xi_{\text{test}}$ are drawn from $\ccalD$, this means that they are feasible by assumption (see Section \ref{sec:labeling}). This equivalently means that if the symbolic task planning algorithm is complete, any failures of Alg. \ref{alg1:cp_ltl_planner} in finding a correct plan are not attributed to the symbolic planner. 
Under this setting, the following three cases may occur as the robot designs its plan.
Case I: We have that $|\mathcal{C}({\ell}_{\text{test}}(t))|=1$, $\forall t\in\{1,\dots,H_{\text{test}}\}$ 
and $\tau_{\phi,\text{test}}\in\ccalC(\bar{\ell}_{\text{test}})$ where $\tau_{\phi,\text{test}}$ is the ground truth plan and \textcolor{black}{prediction set} $\ccalC(\bar{\ell}_{\text{test}})$ is defined as in \eqref{eq:causalCP}.
In this case, the robot will select the correct plan.
Case II: We have that $|\mathcal{C}({\ell}_{\text{test}}(t))|>1$ 
for at least one time step $t\in\{1,\dots,H_{\text{test}}\}$ 
and $\tau_{\phi,\text{test}}\in\ccalC(\bar{\ell}_{\text{test}})$. 
In this case, the robot will select the correct plan \textcolor{black}{assuming users who faithfully provide help.}
%
%
Case III: We have that $\tau_{\phi,\text{test}}\notin\ccalC(\bar{\ell}_{\text{test}})$. The latter means that there exists at least one time step $t$ such that $\tau_{\phi,\text{test}}(t)\notin\ccalC({\ell}_{\text{test}}(t))$.  In this case, the \textcolor{black}{user is unable to provide help and instead halts the operation. Since scenarios drawn from $\ccalD$ admit a unique feasible plan by assumption, the LTL planner also fails to generate alternative sub-tasks. Consequently, in Case III, the mission terminates unsuccessfully.}
%
%
%
%
\textcolor{black}{Notice that we do not know which of these three cases will occur at each step $t$, since the ground truth plan for the test mission scenario is unknown.} However, the probability that either Case I or II will occur is equivalent to the probability $P(\tau_{\phi,\text{test}}\in {\mathcal{C}}(\bar{\ell}_{\text{test}}))$. Due to Proposition \ref{prop:eq_prop} and \eqref{eq:CP2}, we have that $P(\tau_{\phi,\text{test}}\in {\mathcal{C}}(\bar{\ell}_{\text{test}})) \geq 1-\alpha$. Thus, either of Case I and II will occur with probability that is at least equal to $1-\alpha$. Since Cases I-III are mutually and collectively exhaustive, we conclude that the probability that Case III will occur is less than $\alpha$. This means that the probability of HERACLEs generating a correct plan is at least $1-\alpha$.  \textcolor{black}{Due to Assumption \ref{as:A1}, we conclude that the mission success rate is at least $1-\alpha$ completing the proof.} 
\end{proof}

\vspace{-0.2cm}
\section{Experiments}\label{sec:experiments}
\vspace{-0.1cm}


\textcolor{black}{In this section, we demonstrate the performance of the proposed method in various numerical and hardware experiments. 
First, in Section \ref{sec:empirical_validate}, we empirically validate the mission success rate guarantees of HERACLEs (see Theorem \ref{thm1}). We also compare our method against KnowNo \cite{ren2023robots}. We select this baseline as it can also achieve desired mission success rates by letting robots ask for help, using CP, in cases of high uncertainty. 
%
%
Second, in Section \ref{sec:hardware} we provide hardware experiments on real robot platforms to test how the mission completion rate guarantees of HERACLEs may be affected by unmodeled sources of uncertainty (e.g., imperfect robot skills). Demonstrations of HERACLEs are provided in \cite{ltl_llm_video_no_name}. 
Third, in Section \ref{sec:varPred}, we provide examples to compare LTL and LTL-NL formulas in terms of their user-friendliness. Finally, in Section \ref{sec:taskspec}, we  demonstrate how various definitions of the NL-based predicates can affect performance of HERACLEs.
Comparisons against baselines that do not allow robots to ask for help are presented in Appendix \ref{sec:CompI}. 
In all case studies, we pick the following LLMs: GPT-3.5, Llama 2-13b, Llama 3-8b, \textcolor{black}{and Qwen 3-32b}.} 



\subsection{Comparative Numerical Experiments}\label{sec:empirical_validate}

In this section, we empirically validate the mission success rate guarantees of Theorem \ref{thm1} on mobile manipulation tasks. We also provide comparative experiments against KnowNo \cite{ren2023robots}, in terms of help rates required to achieve desired task success rates as well as length of plans and runtimes. \textcolor{black}{Overall, HERACLEs achieves significantly lower help rates due to its symbolic component.} 

\textbf{Environment \& Robot Setup:}  We consider mobile manipulation tasks over an environment populated with $6$ objects: two cans of Coke, one water bottle, one pen, one tin can, and one apple. The water bottle is inside the fridge and the pen is inside a drawer. Thus, grabbing e.g., the pen requires the robot to first open the drawer if it is closed. The status of these containers  (open or closed) is not known a-priori and, therefore, not included in the initial environment description in $\ell(1)$. Instead, it can be provided online through sensor feedback as described in Section \ref{sec:LLMplan}. The action space $\ccalA$ includes $5$ actions as defined in Fig. \ref{fig:prompt}. The action `remain idle' in $\ccalA$ is useful when a sub-task can be accomplished in less than $T$ time steps (see Section \ref{sec:LLMplan}). Given a prompt $\ell(t)$, the number of choices $s$ that the LLM can pick from is $|\ccalS|=17$. Recall that this set is constructed using $\ccalA$ and all objects/locations in the environment where the actions in $\ccalA$ can be applied. We select $T=7$ for the LLM-based action planner.  

\textbf{LTL-NL Tasks:} We construct a distribution $\mathcal{D}$ to sample LTL-NL formulas following the approach in \cite{ren2023robots}. For all generated formulas, there exists only one sequence of sub-tasks that can lead to mission completion as required by Th. \ref{thm1} and Appendix \ref{sec:RelaxI}; this is relaxed in Section \ref{sec:hardware}. Specifically, $\mathcal{D}$ is designed to generate scenarios $\xi$ of `easy,' `medium,' and `hard' difficulty, with probabilities $0.2$, $0.1$, and $0.7$, respectively. \textcolor{black}{Each difficulty category comprises hundreds of LTL-NL formulas, with the difficulty level determined by the complexity of the formulas in terms of the number of NL-based predicates and temporal/logical operators. We note that this may not be an accurate classification of the mission scenarios as task complexity may depend on additional parameters (e.g.,  number of actions required for completion of each predicate/sub-task or whether the sub-tasks requiring opening containers to retrieve items) while also what makes a task complex may really depend on the employed LLM.}
When a difficulty category is chosen, a random LTL-NL formula is selected from the corresponding predefined set. We emphasize that  $\ccalD$ is not known to our framework. More complex distributions $\ccalD$ can also be constructed as CP, and, consequently, HERACLEs, is distribution-free. 
%
%

We sample $111$ scenarios from $\ccalD$ and group them based on their difficulty category.  \textit{Easy} tasks include $25$ LTL-NL formulas of the form $\phi=\Diamond \pi_1$ where $\pi_1$ is defined as `Move object $e$ to location $\bbx$' for various objects $e$ and locations $\bbx$; \textit{Medium} tasks include $15$ LTL-NL formulas defined as $\phi=\Diamond \pi_1 \wedge (\neg \pi_1 \ccalU \pi_2)$, such a  task requires eventually completing the sub-tasks $\pi_1$ and $\pi_2$, but $\pi_2$ must be completed strictly before $\pi_1$. \textit{Hard} tasks include $71$ LTL-NL formulas defined over four to six predicates defined as before. An example of such formula is 
$\phi = (  \neg( \pi_1 \vee \pi_2 \vee \pi_3 \vee \pi_4 ) ) \ccalU
(  \pi_1 \wedge \neg\pi_2 \wedge \neg\pi_3 \wedge \neg\pi_4 \wedge
   \rightarrow(  \neg( \pi_1 \vee \pi_2 \vee \pi_3 \vee \pi_4 )  \ccalU
       ( \pi_2 \wedge \neg\pi_1 \wedge \neg\pi_3 \wedge \neg\pi_4 \wedge
         \rightarrow(  \neg( \pi_1 \vee \pi_2 \vee \pi_3 \vee \pi_4 )  \ccalU
             ( \pi_3 \wedge \neg\pi_1 \wedge \neg\pi_2 \wedge \neg\pi_4 \wedge
               \rightarrow(  \neg( \pi_1 \vee \pi_2 \vee \pi_3 \vee \pi_4 )  \ccalU
                   ( \pi_4 \wedge \neg\pi_1 \wedge \neg\pi_2 \wedge \neg\pi_3 )
               )
             )
         )
       )
   )
)$, which enforces that $\pi_1$, $\pi_2$, $\pi_3$, and $\pi_4$ must be done only once, and in this specific order exclusively.



\textbf{\textcolor{black}{Setting Up the Baseline:}} 
To enable meaningful and fair comparisons against the KnowNo baseline, we have considered the following settings. \textit{First}, since KnowNo uses NL for mission specification, we convert the LTL-NL formulas into NL instructions by translating each temporal/logical operator and atomic proposition into their corresponding NL description. For instance, a medium-level task $\phi=\Diamond \pi_1 \wedge (\neg \pi_1 \ccalU \pi_2)$ can be translated as `Eventually finish \{NL description of $\pi_1$\} and \{NL description of $\pi_2$\}. Do not do \{NL description of $\pi_1$\} until \{NL description of $\pi_2$\} is completed'. The resulting NL instructions serve as an input to KnowNo. \textit{Second} we set up the MCQA-based prompt of KnowNo using the set of decisions/options $\ccalS$ that HERACLEs uses; see also Remark \ref{rem:KnowNo}. \textcolor{black}{Third, we consider fully known environments  (e.g., known obstacles) that do not prevent the robot from accomplishing any sub-task included in $\phi$. As a result, both HERACLEs and KnowNo are applied offline to compute plans. We emphasize that we have considered this setting only to ensure fairness to the baseline. The reason is that our method allows the robot to seek assistance from the symbolic planner in case the environment prevents completion of a sub-task; a feature that does not exist in  KnowNo.
Demonstrations in environments that prevent the accomplishment of certain sub-tasks are presented in \cite{ltl_llm_video_no_name} and discussed in Section \ref{sec:hardware}. 
Fourth, we require KnowNo to complete the plan within $H=T\times K$ steps, where $K$ is the number of predicates in $\phi$ and $T=7$ is the hyper-parameter used in Section \ref{sec:hardware}. We made this choice since HERACLEs also completes a sub-plan for each sub-task in $\phi$ within $T$ steps. 
} 

\begin{table*}[ht]
\centering
\begin{tabular}{|c|c|cc|cc|}
\hline
\textcolor{black}{Model} & \textcolor{black}{Method} 
& \multicolumn{2}{c|}{\textcolor{black}{$1 - \alpha = 0.90$}} 
& \multicolumn{2}{c|}{\textcolor{black}{$1 - \alpha = 0.95$}} \\ \cline{3-6}
 & & \multicolumn{1}{c|}{\textcolor{black}{Success}} & \textcolor{black}{Help} & \multicolumn{1}{c|}{\textcolor{black}{Success}} & \textcolor{black}{Help} \\ \hline

\multirow{2}{*}{\textcolor{black}{Llama 2-13b}} 
  & \textcolor{black}{\textbf{Ours}}    
    & \multicolumn{1}{c|}{\textcolor{black}{93.3\%}} 
    & \textcolor{black}{\textbf{9\%}} 
    & \multicolumn{1}{c|}{\textcolor{black}{97.2\%}} 
    & \textcolor{black}{\textbf{12\%}} \\ \cline{2-6}
  & \textcolor{black}{KnowNo}             
    & \multicolumn{1}{c|}{\textcolor{black}{93.6\%}} 
    & \textcolor{black}{67.2\%} 
    & \multicolumn{1}{c|}{\textcolor{black}{96.7\%}} 
    & \textcolor{black}{78\%} \\ \hline

\multirow{2}{*}{\textcolor{black}{Llama 3-8b}}  
  & \textcolor{black}{\textbf{Ours}}    
    & \multicolumn{1}{c|}{\textcolor{black}{91.0\%}} 
    & \textcolor{black}{\textbf{20\%}}
    & \multicolumn{1}{c|}{\textcolor{black}{98.1\%}} 
    & \textcolor{black}{\textbf{24\%}} \\ \cline{2-6}
  & \textcolor{black}{KnowNo}             
    & \multicolumn{1}{c|}{\textcolor{black}{90.3\%}} 
    & \textcolor{black}{55.8\%} 
    & \multicolumn{1}{c|}{\textcolor{black}{95.1\%}} 
    & \textcolor{black}{64.4\%} \\ \hline

\multirow{2}{*}{\textcolor{black}{Qwen 3-32b}}  
  & \textcolor{black}{\textbf{Ours}}    
    & \multicolumn{1}{c|}{\textcolor{black}{92.9\%}} 
    & \textcolor{black}{\textbf{8.4\%}}
    & \multicolumn{1}{c|}{\textcolor{black}{97.7\%}} 
    & \textcolor{black}{\textbf{10.1\%}} \\ \cline{2-6}
  & \textcolor{black}{KnowNo}             
    & \multicolumn{1}{c|}{\textcolor{black}{90.5\%}} 
    & \textcolor{black}{53.2\%} 
    & \multicolumn{1}{c|}{\textcolor{black}{95.4\%}} 
    & \textcolor{black}{61.6\%} \\ \hline

\end{tabular}
\caption{\textcolor{black}{Comparison of mission success and user help rates for HERACLEs and KnowNo across Llama2-13b, Llama3-8b, and Qwen 3-32b under two confidence levels.}}
\label{tab:heracles_knowno_allblue}
\vspace{-0.8cm}
\end{table*}

\begin{figure}[t] 
\centering
\includegraphics[width=\linewidth]{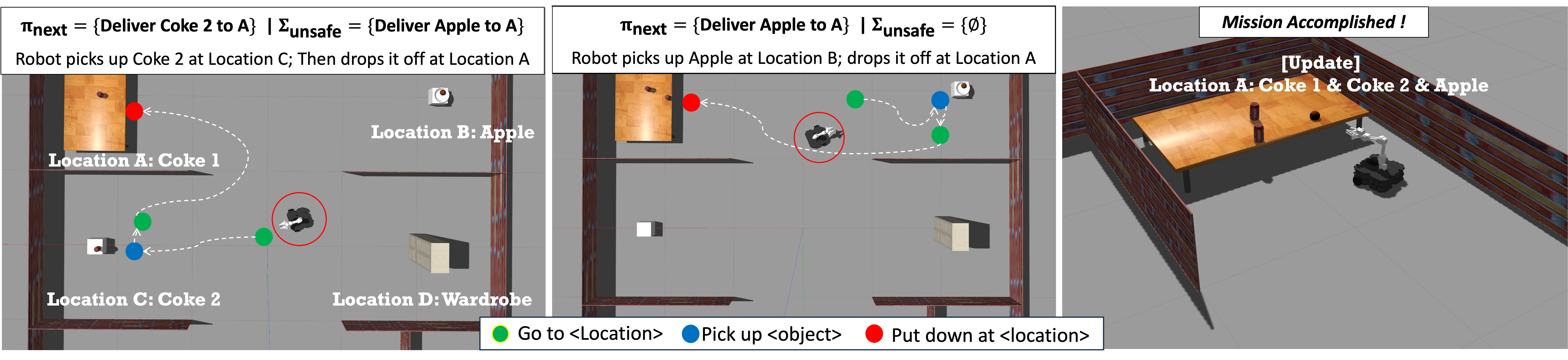}\vspace{-0.4cm} 
\caption{The execution of a robot plan using ROS/Gazebo \cite{gazebo} for the LTL-NL mission: $\phi=\Diamond \pi_1 \wedge (\neg \pi_1 \ccalU \pi_2)$, where $\pi_1$ refers to `Deliver Apple to A' and $\pi_2$ refers to `Deliver Coke 2 to A'. Here $\phi$ requires that the apple should not be delivered to A prior to Coke 2. \textcolor{black}{The left two snapshots illustrate the NL-based sub-tasks that the robot accomplished in the order that they were generated by the symbolic planner.} The corresponding robot plans are also depicted. The right snapshot shows the environment when the task is completed.} \vspace{-0.5cm}
\label{fig:snapshot_case2}
\end{figure}

\textbf{Empirical Mission Success Rates \& Help Rates for HERACLEs:}
\textcolor{black}{We empirically validate the theoretical guarantees for the mission success rate of HERACLEs (Theorem~\ref{thm1}) using Llama 2-13b, Llama 3-8b, \textcolor{black}{and Qwen 3-32b models}. For each validation scenario, we collect $20$ calibration sequences from $\ccalD$. A plan is then computed using Alg.~\ref{alg1:cp_ltl_planner}, while allowing the robot to request help from users when needed. The empirical mission success rate is obtained as follows: among the $111$ generated plans, we compute the ratio of those that are ground truth with respect to their corresponding LTL-NL task $\phi$. This process is repeated $100$ times, and the average ratio is reported as the empirical mission success rate. Table~\ref{tab:heracles_knowno_allblue} summarizes the results for two confidence levels ($1-\alpha = 0.90$ and $1-\alpha = 0.95$). \textcolor{black}{At $1-\alpha = 0.90$, HERACLEs achieves success rates of $93.3\%$, $91.0\%$, and $92.9\%$ with Llama 2-13b, Llama 3-8b, and Qwen 3-32b, respectively.} These validate the theoretical guarantees. \textcolor{black}{The corresponding help rates are $9\%$, $20\%$, and $8.4\%$. At $1-\alpha = 0.95$, the success rates further increase to $97.2\%$, $98.1\%$, and $97.7\%$, respectively, with help rates of $12\%$, $24\%$, and $10.1\%$.} Observe that, as expected, the help rates increase as $1-\alpha$ increases. 
A robot plan synthesized for one of the considered \textit{medium} tasks using HERACLEs and GPT 3.5 is shown in Fig. \ref{fig:snapshot_case2}}


\textbf{Comparisons against KnowNo:} 
We repeated the above process for KnowNo equipped with all three models, using the same validation and calibration data as HERACLEs.
\textcolor{black}{Specifically, with Llama 3-8b, KnowNo achieves $90.25\%$ and $95.12\%$ mission success rates with help rates of $55.8\%$ and $64.4\%$ at $1-\alpha=0.9$ and $1-\alpha=0.95$, respectively. With Llama 2-13b, KnowNo achieves $93.6\%$ and $96.7\%$ success rates, with help rates of $67.2\%$ and $78\%$. \textcolor{black}{With Qwen 3-32b, KnowNo achieves $90.5\%$ and $95.4\%$ success rates, but requires help rates of $53.2\%$ and $61.6\%$, respectively.}
Overall, both methods satisfy the theoretical guarantees on mission success rates. However, KnowNo consistently demands much higher help rates across all models. This highlights the benefit of the symbolic component in HERACLEs, which decomposes the planning problem into smaller sub-problems that the LLM can solve more effectively. In contrast, KnowNo directly relies on the LLM to plan for the entire long-horizon task.
In terms of runtime, both methods are comparable: generating a complete plan takes approximately $70$ seconds with Llama 2-13b, $75$ seconds with Llama 3-8b, \textcolor{black}{and $288$ seconds with Qwen 3-32b}.\footnote{\textcolor{black}{We note that these runtimes are implementation-specific. In our implementation, to design an action $s(t)$, using either method, we query the LLM sequentially, $|\ccalS|$ times, to obtain the confidence score per option. These scores are used to construct $\mathcal{C}(\ell(t))$. These $|\ccalS|$ queries at time $t$ could be conducted in parallel, since they are independent of one another, to accelerate the construction of $\mathcal{C}(\ell(t))$; however, this was not possible in our implementation due to limited memory. Each query required about $0.4$ seconds on average with the Llama models, \textcolor{black}{and about $1.6$ seconds with Qwen 3-32b}. }} The LTL planner in HERACLEs further requires only $2.75 \times 10^{-5}$ seconds on average to generate sub-tasks. Finally, the average length of successful plans across all difficulty levels is $15$ steps for both methods, excluding idle actions.}

\vspace{-0.2cm}
\begin{rem}[LLM model \& Help Rates]
    \textcolor{black}{The quality of the LLM model can critically affect the help rates, with more effective models resulting in lower rates. For instance, the help rate of our method with GPT 3.5 is $2.56\%$, $2.63\%$, and $2.70\%$ for $1-\alpha=0.95$, $0.96$, and $0.97$, respectively.}
\end{rem}

\begin{rem}[Set of Options for KnowNo]\label{rem:KnowNo}
    \textcolor{black}{KnowNo, as originally designed in \cite{ren2023robots}, queries an LLM at each time step to generate a finite set of robot decisions, which are presented as choices in its MCQA-based prompt. These choices include an additional option `other option, not listed' to account for cases where all LLM-generated options are nonsensical.  If the LLM fails to provide valid options, then that would yield an `incomplete' plan ending with `other option, not listed', which would make direct comparisons non-trivial. Thus, we expose KnowNo to the set $S$ of decisions that HERACLES uses, without asking the LLM to generate a set of options.}
\end{rem}

\subsection{Hardware Experiments}\label{sec:hardware}

\textcolor{black}{In this section, we conduct hardware experiments to evaluate robustness of the proposed method against un-modeled sources of uncertainty such as imperfect execution of robot skills.} 

\textcolor{black}{\textbf{Environment \& Robot Setup:} We consider a robot tasked with road-maintenance missions, in a post-disaster city, requiring the robot to block certain streets by placing traffic cones and remove damaged fire hydrants in a temporal/logical order. The environment is populated with $4$ traffic cones and $2$ fire hydrants. Unlike the numerical experiments in Section 
\ref{sec:empirical_validate}, here the geometric structure of the environment is unknown and it may prevent access to certain objects of interest. The action space $\ccalA$ is the same as in Section \ref{sec:empirical_validate} \textcolor{black}{except} the action `$(4, x)$' is removed as there are no containers in the considered environment. This results in a set $\ccalS$ of $13$ decisions.} 

%

\textbf{Empirical Mission Success Rates \& Help Rates:} 
We construct a distribution $\mathcal{D}$ designed to generate tasks of ‘easy’, ‘medium’, and ‘hard’ difficulty levels—similar to those in Section~\ref{sec:empirical_validate}—but which may admit multiple feasible sequences of sub-tasks; see, for example, the LTL formula in Fig.~\ref{fig:env}.
%
We generate $30$ validation scenarios from $\ccalD$ and for each validation scenario, we collect $15$ calibration scenarios.  Using HERACLEs, coupled with Llama 3-8B, we compute robot plans for $1-\alpha=0.9$. The mission success rate is $96.7\%$ with the help rate of $6.3\%$. 

\textcolor{black}{\textbf{Execution of Plans:}
We select $12$ validation scenarios where HERACLEs successfully designed a correct plan and tested them on a ground robot platform. For these experiments, we use a TurtleBot3 Waffle Pi 
equipped with a 4-DOF OpenManipulator-X \cite{emanual_turtlebot}
operating within the Mini-City platform developed at WashU. 
The robot employs existing SLAM algorithms for state estimation and mapping \cite{grisetti2007improved}. In $11$ out of $12$ scenarios, the robot successfully executed the plans without any skill failures. The single failed scenario occurred due to a traffic cone slipping off the robot manipulator. Snapshots of the robot navigating the Mini-City to satisfy the LTL-NL formula used in Example \ref{ex:me}, while also asking for help, are provided in Fig. \ref{fig:env}. Additional demonstrations of our hardware experiments 
are provided in \cite{ltl_llm_video_no_name}. Future work will focus on designing robust planners to address uncertainties encountered during deployment.}

\vspace{-0.4cm}
\subsection{Comparisons of LTL-NL and LTL Formulas}\label{sec:varPred}
\textcolor{black}{
In this section, we provide examples to demonstrate that LTL-NL formulas are more user-friendly than their corresponding LTL formulas defined over the system states $\bbp(t)$ capturing the same mission requirements. As a performance metric, we employ the `length' of the specification, i.e., the total number of atomic predicates and temporal and logical operators. In what follows, we select an LTL-NL formula for each case study considered in Appendix \ref{sec:CompI} and we manually convert it into the respective LTL formula.
%
%
%
First, consider the LTL-NL formula $\phi=\Diamond\pi_1$ from Case Study I, where the NL-based predicate $\pi_1$ is true if the robot delivers a bottle of water to location $A$. This formula only has $1$ temporal/logical operator and  $1$ predicate. Using system-based predicates, the same task will be written where $\pi_1'$ is true if the robot position is close enough to the bottle of water, $\pi_2'$ is true if the robot grabs the bottle successfully, $\pi_3'$ is true if the robot position is close enough to location $A$, and $\pi_4'$ is true if the robot puts down the bottle of water. 
Second, consider the LTL-NL formula $\phi=\Diamond\pi_1\wedge\Diamond\pi_2$ from Case Study II where both $\pi_1$ and $\pi_2$ model delivery tasks as before. This formula has $3$ logical/temporal operators and $2$ predicates. Expressing the same task using system-based predicates as before would result in $15$ logical/temporal operators and $8$ predicates, i.e., $\phi=\Diamond(\pi_1'\wedge(\Diamond\pi_2'\wedge(\Diamond\pi_3'\wedge(\Diamond\pi_4'))))\wedge\Diamond(\pi_5'\wedge(\Diamond\pi_6'\wedge(\Diamond\pi_7'\wedge(\Diamond\pi_8'))))$. Third, we select the following formula from Case Study III: $\phi=\Diamond \pi_1 \wedge \Diamond \pi_2 \wedge \Diamond \pi_3  \wedge \Diamond \pi_4 \wedge (\neg \pi_4 \ccalU \pi_1)$. This LTL-NL formula has $10$ temporal/logical operators and $4$ predicates, whereas the corresponding LTL formula over system-based predicates requires $46$ temporal/logical operators and $16$ predicates. Observe that the difference in the length of LTL and LTL-NL formulas becomes more pronounced as the complexity of the mission requirements increases.}

\vspace{-0.2cm}
\subsection{Effect of Predicate Definitions on Planning Performance}\label{sec:taskspec}
\vspace{-0.1cm}
\textcolor{black}{
In this section, first we show that a fixed mission can be represented by various LTL-NL formulas that differ in the definition of the NL-based predicates. Then, we demonstrate how various definitions of these predicates, resulting in equivalent LTL-NL formulas, can affect performance of HERACLEs.}

Specifically, we consider a simple mission requiring to deliver objects A, B, C, and D to desired destinations in any order. 
%
We can express this task using an LTL-NL formula with four APs: $\phi_1=\Diamond\pi_A \wedge \Diamond\pi_B \wedge \Diamond\pi_C \wedge \Diamond\pi_D$. In $\phi_1$, $\pi_i$ is true when the object $i\in\{A,B,C, D\}$ is delivered to the intended destination. 
An equivalent LTL-NL formula is $\phi_2=\Diamond\pi_{A,B} \wedge \Diamond\pi_C \wedge \Diamond\pi_D$ where the only difference is that now we have defined a predicate $\pi_{A,B}$ that is true if both $A$ and $B$ are delivered to their destinations in any order. In a similar way, we can define the equivalent LTL-NL formulas $\phi_3=\Diamond\pi_{A,B,C} \wedge \Diamond\pi_D$ and $\phi_4=\Diamond\pi_{A,B,C,D}$. Observe that formula $\phi_4$ is the most `user-friendly' one as it has the fewest predicates and temporal operators. However, this results in increasing the number of sub-tasks that each predicate models. This in turn requires the LLM to reason about more complex task requirements that would otherwise have been handled by the symbolic planner. For instance, in $\phi_4$, the symbolic planner generates the subtask $\pi_{A,B,C,D}$ and the LLM needs to design a feasible plan for it. Increasing the complexity of the sub-tasks modeled by the predicates may in turn result in higher help rates. Specifically, we run HERACLEs using Llama 2-13b model to compute plans for each LTL-NL formula. When $1-\alpha=0.9$, the average help rates over $10$ trials for $\phi_1$, $\phi_2$, $\phi_3$, and $\phi_4$ were $10\%$, $35\%$, $50\%$, and $85\%$ respectively; each trial differs in the calibration dataset. Notice that for $\phi_4$, HERACLEs is essentially equivalent to \cite{ren2023robots} since the symbolic planner cannot perform task decomposition by construction of $\phi_4$.

\vspace{-0.3cm}
\section{Conclusions, Limitations,  and Future Work}
\vspace{-0.1cm}
%
In this paper, we proposed HERACLEs, a new neuro-symbolic planner for LTL-NL tasks. We showed, both theoretically and empirically, that it can achieve desired mission success rates due to a conformal interface between the symbolic planner and the LLM. We also provided comparative experiments showing that it outperforms LLM-based planners in terms of planning accuracy and help rates as well as examples to demonstrate its user-friendliness compared to symbolic planners. Future work will address imperfect skill execution and out-of-distribution mission scenarios.
\vspace{-0.2cm}

\bibliographystyle{ACM-Reference-Format}
\bibliography{YK_bib}

\appendix
\vspace{-0.3cm}
\begin{appendices}

\section{Comparisons against Non-Conformalized Neural Planners}\label{sec:CompI}

\textcolor{black}{In this Appendix, we compare HERACLEs against baselines \cite{ahn2022can,chen2023scalable,singh2023progprompt}  which, however, do not utilize CP and do not allow robots to ask for help. Our comparative experiments show that the ability of these approaches to design correct plans tends to decrease as the number of temporal and logical requirements in the mission increases. On the contrary, the planning performance of HERACLEs consistently outperforms these baselines (even if the help mode is deactivated). We consider the same environment and robot setup as in Section \ref{sec:empirical_validate}.}

\textbf{Setting up the Baselines:} As the baselines for our experiments, we employ \textcolor{black}{three} state-of-the-art LLM-based planners: SayCan \cite{ahn2022can}, CMAS \cite{chen2023scalable}, \textcolor{black}{and ProgPrompt\cite{singh2023progprompt}}. 
Unlike HERACLEs, these baselines require the overall mission to be fully described using NL. Thus, we manually convert LTL-NL tasks into NL ones, which are then used as inputs for the baselines; see Section \ref{sec:empirical_validate}. SayCan generates decisions, taking into consideration the likelihood that the corresponding action will be executed correctly. Since we consider perfect execution of robot actions, we assume that the affordance functions in \cite{ahn2022can} always return a value of $1$ for all decisions. 
We note that SayCan converts the NL mission planning problem into a sequence of MCQA problems, as HERACLEs does too. Then SayCan selects the action with the highest softmax score: $s(t)=\arg\max_{s\in\mathcal{S}} g(s|\ell(t))$ for every MCQA problem.
%
%
We also compare our method against the CMAS planner \cite{chen2023scalable}. CMAS is originally developed for teams of $N$ robots. We applied it in our setting using $N=1$. Note that \textcolor{black}{CMAS and ProgPrompt} do not employ the MCQA setup (unlike HERACLEs and SayCan). Instead, at every time step, the LLM generates new tokens corresponding to robot actions. 
\textcolor{black}{ProgPrompt - unlike our method, Saycan, and CMAS - requires its prompt to be written in the format of an executable programming language such as Python. 
}
%

\textcolor{black}{To make our comparisons fair, we have enforced the following requirements: (i) All methods are exposed to the same set of actions $\ccalS$. (ii) All methods \textcolor{black}{(except ProgPrompt)} share the same prompt structure. The only difference is that part (C) in the prompts for HERACLEs includes the description of the sub-task generated by the symbolic planner while part (C) for SayCan, CMAS, and ProgPrompt includes the overall mission expressed in NL (as these baselines do not consider task decomposition). \textcolor{black}{The main body of the prompt in ProgPrompt is described in Python-like format as in the original work \cite{singh2023progprompt}.} (iii) All actions are executed perfectly.
(iv) \textit{We completely deactivate the help mode from our planner since this capability does not exist in these baselines.} We emphasize that we have applied (iv) only to make comparisons fair against the baselines as otherwise our planner can outperform them by picking a low enough value for $\alpha$ and asking for help from users. This choice also allows us to assess the `nominal' performance of our planner when help from users is not available.
%
The requirement (iv) means that we remove altogether the CP component from our planner. This implies that HERACLEs always picks the decision $s(t)=\arg\max_{s\in\mathcal{S}} g(s|\ell(t))$ (as SayCan does too). Also, to accommodate (iv), we consider environments with known obstacles that do not prevent the robot from accomplishing any sub-task included in $\phi$; see Section \ref{sec:empirical_validate}. This implies that HERACLEs will not trigger an assistance request from the symbolic planner. Again, we note that we have applied (iv) only to ensure fairness to other baselines that do not have the capability of asking for help from users or symbolic planners. 
(v) We require all methods to complete the plan within $H=T\times K$ steps, where $K$ is the number of predicates in $\phi$ and $T=7$ is the hyper-parameter used in Section \ref{sec:LLMplan}.} 
In what follows, we report the performance of our method and the baselines over the LTL-NL formulas of various levels of difficulty. As performance metrics, we use runtimes as well as planning accuracy defined as the percentage of scenarios where a planner generates a correct plan. The results are also summarized in Table \ref{table:acc_runtime_comp}.

\begin{table}[t]
\centering
\begin{tabular}{|c|c|ccc|c|}
\hline
\multirow{2}{*}{\textcolor{black}{Model}} & \multirow{2}{*}{\textcolor{black}{Method}} & \multicolumn{3}{c|}{\textcolor{black}{Case Study}} & \multirow{2}{*}{\textcolor{black}{Avg. Runtime (s)}} \\ \cline{3-5}
                       &                         & \multicolumn{1}{c|}{\textcolor{black}{I (Easy)}} & \multicolumn{1}{c|}{\textcolor{black}{II (Medium)}} & \textcolor{black}{III (Hard)} &                                 \\ \hline
\multirow{4}{*}{\textcolor{black}{GPT 3.5}}     
  & \textcolor{black}{\textbf{Ours}}      
    & \multicolumn{1}{c|}{\textcolor{black}{\textbf{96\%}}}   
    & \multicolumn{1}{c|}{\textcolor{black}{\textbf{93.3\%}}}   
    & \textcolor{black}{\textbf{93\%}}     
    & \textcolor{black}{\textbf{14}}                              \\ \cline{2-6}
  & \textcolor{black}{SayCan}               
    & \multicolumn{1}{c|}{\textcolor{black}{$96\%$}}   
    & \multicolumn{1}{c|}{\textcolor{black}{$40\%$}}      
    & \textcolor{black}{$14.08\%$}  
    & \textcolor{black}{15}                              \\ \cline{2-6}
  & \textcolor{black}{CMAS}           
    & \multicolumn{1}{c|}{\textcolor{black}{$56\%$}}   
    & \multicolumn{1}{c|}{\textcolor{black}{$20\%$}}      
    & \textcolor{black}{$0\%$}      
    & \textcolor{black}{3}                               \\ \cline{2-6}
  & \textcolor{black}{ProgPrompt}  
    & \multicolumn{1}{c|}{\textcolor{black}{48\%}}   
    & \multicolumn{1}{c|}{\textcolor{black}{16\%}}      
    & \textcolor{black}{2.5\%}    
    & \textcolor{black}{4}                               \\ \hline\hline
\multirow{4}{*}{\textcolor{black}{Llama 2-13b}} 
  & \textcolor{black}{\textbf{Ours}}      
    & \multicolumn{1}{c|}{\textcolor{black}{\textbf{86\%}}}   
    & \multicolumn{1}{c|}{\textcolor{black}{\textbf{81\%}}}      
    & \textcolor{black}{\textbf{80\%}}     
    & \textcolor{black}{\textbf{71}}                              \\ \cline{2-6}
  & \textcolor{black}{SayCan}               
    & \multicolumn{1}{c|}{\textcolor{black}{$86\%$}}   
    & \multicolumn{1}{c|}{\textcolor{black}{$18\%$}}      
    & \textcolor{black}{$10\%$}     
    & \textcolor{black}{75}                              \\ \cline{2-6}
  & \textcolor{black}{CMAS}           
    & \multicolumn{1}{c|}{\textcolor{black}{$53\%$}}   
    & \multicolumn{1}{c|}{\textcolor{black}{$9\%$}}       
    & \textcolor{black}{$0\%$}      
    & \textcolor{black}{15}                              \\ \cline{2-6}
  & \textcolor{black}{ProgPrompt}
    & \multicolumn{1}{c|}{\textcolor{black}{46.7\%}} 
    & \multicolumn{1}{c|}{\textcolor{black}{9\%}}    
    & \textcolor{black}{0\%}      
    & \textcolor{black}{18}                              \\ \hline\hline
\multirow{4}{*}{\textcolor{black}{Llama 3-8b}}  
  & \textcolor{black}{\textbf{Ours}}       
    & \multicolumn{1}{c|}{\textcolor{black}{\textbf{87.5\%}}} 
    & \multicolumn{1}{c|}{\textcolor{black}{\textbf{90.9\%}}}     
    & \textcolor{black}{\textbf{90\%}}    
    & \textcolor{black}{\textbf{76}}                              \\ \cline{2-6}
  & \textcolor{black}{SayCan}               
    & \multicolumn{1}{c|}{\textcolor{black}{$87.5\%$}} 
    & \multicolumn{1}{c|}{\textcolor{black}{$36\%$}}      
    & \textcolor{black}{$18\%$}    
    & \textcolor{black}{78}                              \\ \cline{2-6}
  & \textcolor{black}{CMAS}           
    & \multicolumn{1}{c|}{\textcolor{black}{$50\%$}}   
    & \multicolumn{1}{c|}{\textcolor{black}{$27\%$}}      
    & \textcolor{black}{$18\%$}    
    & \textcolor{black}{17}                              \\ \cline{2-6}
  & \textcolor{black}{ProgPrompt}
    & \multicolumn{1}{c|}{\textcolor{black}{43.8\%}} 
    & \multicolumn{1}{c|}{\textcolor{black}{18.1\%}}     
    & \textcolor{black}{0\%}  
    & \textcolor{black}{25}                              \\ \hline\hline
\multirow{4}{*}{\textcolor{black}{Qwen 3-32b}}  
  & \textcolor{black}{\textbf{Ours}}       
    & \multicolumn{1}{c|}{\textcolor{black}{\textbf{ 96\%}}} 
    & \multicolumn{1}{c|}{\textcolor{black}{\textbf{ 93.3\%}}}     
    & \textcolor{black}{\textbf{95\%}}    
    & \textcolor{black}{\textbf{300}}                              \\ \cline{2-6}
  & \textcolor{black}{SayCan}               
    & \multicolumn{1}{c|}{\textcolor{black}{96\%}} 
    & \multicolumn{1}{c|}{\textcolor{black}{63\%}}      
    & \textcolor{black}{25\%}    
    & \textcolor{black}{350}                              \\ \cline{2-6}
  & \textcolor{black}{CMAS}           
    & \multicolumn{1}{c|}{\textcolor{black}{80\%}}   
    & \multicolumn{1}{c|}{\textcolor{black}{48\%}}      
    & \textcolor{black}{30\%}    
    & \textcolor{black}{72}                              \\ \cline{2-6}
  & \textcolor{black}{ProgPrompt}
    & \multicolumn{1}{c|}{\textcolor{black}{96\%}} 
    & \multicolumn{1}{c|}{\textcolor{black}{93\%}}     
    & \textcolor{black}{42\%}  
    & \textcolor{black}{90}                              \\ \hline
\end{tabular}
\caption{\textcolor{black}{Comparison of planning accuracy and runtime for HERACLES (CP/help mode deactivated) against non-conformalized baselines, using GPT 3.5, Llama 2-13b, Llama 3-8b, and Qwen 3-32b.}}
\label{table:acc_runtime_comp}
\vspace{-1cm}
\end{table}
\normalsize

\textbf{Case Study I (Easy):} We consider $25$ LTL-NL formulas of the form $\phi=\Diamond \pi_1$ where $\pi_1$ is defined as `Move object $e$ to location $\bbx$' for various objects $e$ and locations $\bbx$.
Using GPT 3.5, the accuracy of HERACLEs was $96\%$ while the accuracy of SayCan, CMAS, and ProgPrompt were $96\%$, $56\%$, and $48\%$, respectively. Using Llama 2-13b, the accuracy of our method and SayCan dropped to 86\%, while for CMAS and ProgPrompt, it dropped to 53\% and 46.7\%, respectively. 
\textcolor{black}{Using Qwen 3-32b, HERACLEs achieved 96\% of accuracy, while SayCan, CMAS, and ProgPrompt achieved 96\%, 80\%, and 96\%, respectively.}
%
Notice that the performance of our method and SayCan is the same since the mission $\phi$ cannot be further decomposed, and both methods share the same prompt as well as the same action selection mechanism when the help mode is removed from our method.

\textbf{Case Study II (Medium):} We consider $15$ LTL-NL formulas defined as either $\phi_1=\Diamond \pi_1 \wedge \Diamond \pi_2$ or $\phi_2=\Diamond \pi_1 \wedge (\neg \pi_1 \ccalU \pi_2)$. The task $\phi_1$ requires to eventually complete $\pi_1$ and $\pi_2$ in any order while $\phi_2$ requires $\pi_2$ to be completed strictly before $\pi_1$. The APs $\pi_1$ and $\pi_2$ are defined as before. 
%
Using GPT 3.5, the accuracy of our planner, SayCan, CMAS, and ProgPrompt are $93.3\%$, $40\%$, $20\%$, and $16\%$, respectively. \textcolor{black}{Similar trends are observed with Llama 2-13b and Qwen 3-32b; see Table \ref{tab:heracles_knowno_allblue}. Overall, each method achieved its best performance when paired with Qwen 3-32b.}
%

\textbf{Case Study III (Hard):} We consider $71$ LTL-NL formulas defined over four to six predicates such as $\phi=\Diamond \pi_1 \wedge \Diamond \pi_2 \wedge \Diamond \pi_3 \wedge (\neg \pi_3 \ccalU \pi_2) \wedge \Diamond \pi_5 \wedge (\neg \pi_2 \ccalU \pi_5) \wedge (\neg \pi_5 \ccalU \pi_1) \wedge \Diamond \pi_4$.
The accuracy of our planner, SayCan, CMAS, and ProgPrompt is $93\%$, $14.08\%$, $0\%$, and $2.5\%$, respectively, using GPT 3.5. \textcolor{black}{Similar trends are observed when Llama 2-13b and Qwen 3-32b are used; see Table \ref{tab:heracles_knowno_allblue}. As in Case Study II,  each method achieved its best performance when paired with Qwen 3-32b.}

\textbf{Runtimes:} 
In the above case studies, the average runtime to generate the entire plan for different methods using GPT 3.5, Llama 2-13b, Llama 3-8b, \textcolor{black}{and Qwen 3-32b} are reported in Table \ref{table:acc_runtime_comp}. Observe that HERACLEs and SayCan have comparable runtimes as they both rely on the MCQA framework for action selection, requiring $|\ccalS|$ LLM queries to select $s(t)$. In contrast, CMAS and ProgPrompt require less runtime across all models because they need only one LLM query at each time $t$ to design $s(t)$. Additionally, note that the runtimes for all methods are significantly higher when using \textcolor{black}{Llama or Qwen} models compared to GPT-3.5. This is primarily because Llama models are stored and run on our local computers, whereas GPT-3.5 is accessed from OpenAI servers.

\textbf{Summary of Comparisons:} \textcolor{black}{Observe that the performance gap, in terms of planning accuracy, between HERACLEs and the baselines increases significantly as the task complexity increases \textcolor{black}{for all considered LLMs}. Also, notice that the performance of HERACLEs does not change significantly across the considered case studies. The reason is that HERACLEs decomposes the overall planning problem into smaller ones that can be handled efficiently by the LLM. This is in contrast to SayCan, CMAS, and ProgPrompt, where the LLM is responsible for generating plans directly for the original long-horizon task. Additional comparisons showing that the ability of LLMs to design correct plans deteriorates as temporal and logical requirements are incorporated into the mission can be found in \cite{chen2023autotamp}.}  We observe SayCan’s superior performance over CMAS and ProgPrompt and attribute it to MCQA’s hallucination mitigation. By comparison, CMAS and ProgPrompt generate fresh tokens for planning, making them more prone to hallucinations.

\vspace{-0.3cm}
\section{\textcolor{black}{Mission Scenarios with Multiple Feasible Plans}}\label{sec:RelaxI}

\textcolor{black}{\textbf{CP Analysis:} We begin by discussing how the CP analysis in Section~\ref{sec:labeling} is adapted in case $\ccalD$ generates scenarios admitting multiple feasible plans. First, we consider scenarios where an LTL-NL formula can be satisfied by completing a fixed sequence of sub-tasks, but each sub-task may admit multiple feasible plans; e.g., a sub-task might involve relocating at least one out of $K$ objects. In this case, the key modification lies in the construction of the calibration dataset. Specifically, for each calibration mission scenario, we construct the ground-truth plan for each sub-task by selecting, at each step, the feasible and executable decision with the highest LLM confidence score, given the current environment information. CP can then be applied as usual, yielding prediction sets containing with user-specified probability, the plan $\tau_{\phi,\text{test}}$ consisting of the decisions with the highest LLM confidence scores among all feasible decisions. This can also be formally shown by following the same steps as in Appendices A3-A4 of \cite{ren2023robots}.
Second, we address scenarios where LTL-NL formulas can be satisfied through more than one sequence of sub-tasks and each sequence may admit multiple feasible plans. In such cases, the key modification compared to the above lies in how the symbolic planner selects the next sub-task when queried. Specifically, when the symbolic planner announces a sub-task, among all available sub-tasks (if more than one), it should select the one for which the first feasible action with the highest LLM score towards completing that sub-task has a higher LLM score than the corresponding actions for all other sub-tasks. We acknowledge that this condition is challenging to enforce in symbolic planners; if it is not satisfied, the mission success guarantees may be compromised.}

\textcolor{black}{\textbf{Help Mode \& Theorem \ref{thm1}:}
Second, we discuss how help should be provided during validation scenarios so that Theorem \ref{thm1} holds. Specifically, in case of multiple feasible solutions, the proof of Th. \ref{thm1} follows the same steps. The only difference being that the correct plan $\tau_{\phi,\text{test}}$ refers to the feasible plan comprising the decisions with the highest LLM confidence scores at each step (see the above CP analysis). This has three implications. First, when user help is needed, the user selects the feasible action with the highest LLM score from the prediction set. Thus, the robot must return to the user the scores for each action in the set. Second, when the symbolic planner announces a sub-task, among all available sub-tasks, it should select the one as discussed above.
Third, if at any step $t$ we fall into Case III, i.e., $\tau_{\tau,\text{test}}(t) \notin \mathcal{C}(\ell_{\text{test}}(t))$ and $|\mathcal{C}(\ell_{\text{test}}(t))| > 1$, the user is still unable to provide help, but the LTL planner may generate alternative sub-tasks. This capability can significantly boost the mission success rate beyond $1 - \alpha$ compared to setups where help from the LTL planner is not solicited in Case III.
}

\vspace{-0.3cm}
\section{\textcolor{black}{Relaxing the Assumption of Error-Free Skills}}  \label{sec:relaxPerfectSkills}

\textcolor{black}{Assumption \ref{as:A1} requires perfect execution of robot skills and is required to provide the mission completion guarantees discussed in Thm. \ref{thm1}. Relaxing this assumption necessitates integrating skill imperfections in CP. This requires score functions that estimate a skill’s success probability from the current state. A potential approach is by learning affordance functions\cite{ahn2022can}. 
Then we multiply the skill score by the heuristic LLM uncertainty score (the model's confidence).
A higher product score indicates a greater likelihood that the corresponding action is correct and will be executed successfully. We then apply CP as in Section \ref{sec:labeling} to construct prediction sets. The challenge here is that non-singleton sets can arise due to uncertainties of either the LLM or the robot skills. Help from a user can come in two ways depending on the source of uncertainty.
A potential approach is to decompose the prediction set into LLM-uncertainty and skill-uncertainty subsets.
If the LLM is the source of uncertainty, assistance can be provided by asking a human to select the correct decision. If uncertainty arises from robot skills, assistance may involve the physical execution of decisions by a human (or other robot teammates).
Formally relaxing this assumption and extending Thm. \ref{thm1} to account for imperfect robot skills is part of our future work.} 
\end{appendices}

\end{document}